\documentclass[11pt]{article}
\usepackage[margin=1in]{geometry}
\usepackage{amsmath,amssymb,amsfonts}
\usepackage{graphicx}
\usepackage{booktabs}
\usepackage{microtype}
\usepackage[numbers,sort&compress]{natbib}
\usepackage{xcolor}
\usepackage[colorlinks=true,linkcolor=blue!60!black,citecolor=blue!60!black,urlcolor=blue!60!black]{hyperref}
\usepackage{caption}
\usepackage{needspace}
\usepackage{placeins}
\graphicspath{{figures/}}

\newcommand{\ak}{\textsc{Alpha-Kitten}}
\newcommand{\champ}{\texttt{B-grpo\_grpo\_g870000}}

\title{\ak{}: League Self-Play Beats Open-Source LLMs \\
at a Stochastic Imperfect-Information Card Game}

\author{
  Karthik Ragunath Ananda Kumar \\
  Tavus Inc. \\
  \texttt{a.karthikragunath@gmail.com}\thanks{\raggedright Code, training and evaluation
  scripts, and the paper source are available at
  \url{https://github.com/Karthik-Ragunath/alpha-kitten}, and the trained
  checkpoints at \url{https://huggingface.co/KarthikRagunathAnandaKumar/alpha-kitten}.}
}
\date{}

\begin{document}
\maketitle

\begin{abstract}
We train league self-play agents for Exploding Kittens, a two-player card game
with hidden information, 186 discrete actions and a $\pm1$ reward only at the
end of each game. Ten training runs cover five recipes, and we evaluate the
agents over 1.33 million games. Two open-weight LLMs that see exactly what the
agents see lose 8.5:1 (Qwen3-30B-A3B) and 6.5:1 (Gemma-4-31B). We then study
three parts of the recipe. Keeping fixed outside opponents in training makes a
recipe the hardest to beat for every outside opponent we tested, but its agents
average 17.7th of 20 in a playoff against the other trained agents. Distilling
five different algorithms and a rule-based bot into a six-expert mixture gives
a student weaker than its best teacher, and after self-play, agents distilled
from six checkpoints of one algorithm beat those distilled from the varied
teachers. The mixture-of-experts gate picks different experts in different
game phases, but using one expert instead of two changes only 1.3--2.7\% of
moves. Finally, the best agent depends on how wins are counted: in a 50-agent
round robin, one agent is first by games won but eighth by opponents beaten,
and in 15\% of the trios drawn from the top 20, the three agents beat each
other in a circle.
\end{abstract}

\section{Introduction}
Exploding Kittens is a short card game with a lot of luck in it. One Exploding
Kitten sits somewhere in a shuffled deck, and a player who draws it without a
Defuse loses. Skill still matters: when to draw, when to pass turns to the
opponent, and when to cancel the opponent's move with a Nope. But neither
player can see the other's hand, and a game gives only a win or a loss at the
end. This makes it a hard setting for reinforcement learning, and a fair test
for language models, which can be given the exact same information as a
trained agent.

We train five specialist agents with different algorithms (PPO, GRPO, DPO, SAC
and DREAM), copy them into one mixture-of-experts agent, and then improve that
agent by league self-play. Ten training runs cover five recipes, each changing
one part of this process: which teachers are copied, and which opponents are
played. We evaluate the resulting agents over 1.33 million games: against each
other, against two fixed baseline players, and against two open-weight LLMs.

Self-play populations, from AlphaZero~\citep{silver2018alphazero} onward, are
usually summarised by a single rating~\citep{elo1978rating,bradley1952rank},
and the top-rated checkpoint is the one that gets used. We do the same, and
then check the result in several other ways.

\needspace{6\baselineskip}
\noindent Our main contributions are:
\begin{enumerate}
\item \textbf{A fair test of LLMs against the trained agents}: two LLMs from
different model families play the game seeing exactly what the agents see,
translated into plain text, and choosing from the same list of allowed moves.
Neither side can see the opponent's hand or the order of the deck, and every LLM reply could be read as a move; only 21 of Gemma's 135{,}319
moves were illegal and had to be replaced (Section~\ref{sec:llm}).
\item \textbf{Who you practise against decides who you beat}: changing one
setting, how often an agent trains against fixed outside opponents instead of
copies of itself, made it the best recipe against four very different
outsiders (a random player, a rule-based bot and two LLMs) but one of the
weakest against the other trained agents (Section~\ref{sec:ood}).
\item \textbf{More teachers did not make a better student}: an agent that
copied six teachers (five different algorithms and a rule-based bot) before
self-play ended up weaker than the best of those teachers (measured right
after the distillation stage). After self-play,
agents that had copied six versions of a single algorithm beat agents that had
copied the six different teachers (Section~\ref{sec:distill}).
\item \textbf{How you score decides who wins}: ranking agents by total games
won and by number of opponents beaten crowns different agents, and the
training recipe decides which kind of agent you get, one that crushes weak
opponents or one that narrowly beats everyone (Section~\ref{sec:eval}).
\item \textbf{LLMs know a key rule but rarely use it}: trained agents use the
strongest move in the game (passing an attack back with two extra turns) in
31.9\% of their attacks; Qwen3-30B does so in only 0.6\% and Gemma-4-31B in
3.7\%, even though their instructions state the rule word for word (Section~\ref{sec:llm}).
\end{enumerate}

\section{Related Work}
Most of the questions in this paper already have partial answers in prior
work, so we list what is new here and what is not.

\paragraph{Setting:} Self-play in imperfect-information card games is well
studied: regret minimisation and its deep variants are the standard for
poker~\citep{brown2019deepcfr,steinberger2020dream}, and DouDiZhu, Big~2 and
Hanabi all have large self-play literatures. Applying self-play RL to a
luck-heavy card game is not new. Our contribution is the set of ablations
around it.

\paragraph{Opponent composition:} Self-play is known to overfit to its own
opponents, and prioritised fictitious self-play~\citep{vinyals2019alphastar}
was introduced to widen the pool.  We did not find a controlled ablation
that varies the fixed-opponent ratio across matched runs and reports
in-distribution and out-of-distribution strength separately against several independent out-of-distribution opponent classes (Section~\ref{sec:ood}).

\paragraph{Multi-teacher distillation:} Policy
distillation~\citep{rusu2015policy,hinton2015distilling} and multi-teacher
variants already exist, including for RL policies in game-theoretic
settings~\citep{mepdppo2025}. We add a negative result in a new setting
(Section~\ref{sec:distill}). It matches recent LLM work: \citet{openmopd2026} report that a multi-teacher student, even with each
prompt routed to the right teacher, fell short of its single-domain teachers,
which matches what we see.

\paragraph{Mixture-of-experts policies:} Sparse MoE
layers~\citep{shazeer2017outrageously,fedus2022switch} are standard. As far as
we know, nobody has combined them with self-play RL in a luck-heavy card game,
so we do not claim it as a contribution. We measure
whether the gate's routing changes the agent's moves at all
(Section~\ref{sec:moe}).

\paragraph{Evaluation:} Our ranking results are a measured example of a known
theoretical result: \citet{balduzzi2018reevaluating} showed that a
single rating such as Elo cannot represent players who beat each other in a
circle (A beats B, B beats C, C beats A), and proposed Nash averaging instead.
We add a measurement at scale, and show that the training recipe decides
where an agent sits in those circles.

\section{Method}\label{sec:method}
\subsection{Environment}\label{sec:env}
We use two-player Exploding Kittens, base game only (the official 2022 rules,
with no expansion cards), implemented in PettingZoo~\citep{terry2021pettingzoo}.
The game uses 51 cards: each player is dealt 8 (including one Defuse), 35 go to
the draw pile, and the pile holds exactly one Exploding Kitten. Players take
turns playing action cards (Attack, Skip, Favor, Shuffle, See the Future, Nope,
cat pairs) and then draw; a player who draws the Exploding Kitten without a
Defuse loses. Neither player can see the other's hand or the order of the deck.

The observation is a 512-number vector: the player's own hand, how many of each
card have been discarded, the opponent's hand size, the deck size, the game
phase, how many turns are owed, and any cards seen with See the Future. There
are 186 discrete actions, and a mask marks which of them are legal in the
current state. A game that reaches the turn limit is a draw; with the base-game
rules this almost never happens.

\subsection{Rewards and training signal}\label{sec:reward}
The game gives a reward of $+1$ to the winner and $-1$ to the loser at the end
of the game, $0$ for a draw, and $0$ on every move before that. We use no reward
shaping. Each stage of training turns this end-of-game result into something it
can learn from in a different way (Table~\ref{tab:signal}). Stage~1 of Phase~2
uses no reward at all: it only copies the teachers.

\begin{table}[h]\centering\small
\begin{tabular}{lll}
\toprule
Stage & Algorithm & What it learns from \\
\midrule
Phase 1 & PPO & the $\pm1$ result, spread back over the moves by the critic (GAE, Eq.~\ref{eq:ppo-adv}) \\
Phase 1 & GRPO & the $\pm1$ result of each game, compared with its group of 16 (Eq.~\ref{eq:grpo-adv}) \\
Phase 1 & DPO & pairs of its own won and lost games, not the reward itself \\
Phase 1 & SAC & the $\pm1$ result plus an entropy bonus, inside its $Q$-values \\
Phase 1 & DREAM & regrets estimated from the discounted $\pm1$ result \\
Phase 2, Stage 1 & distillation & no reward: the teachers' move probabilities (Eq.~\ref{eq:distill}) \\
Phase 2, Stage 2 & GRPO & the $\pm1$ result of each game, compared with its block of 64 \\
Phase 2, Stage 2 & PPO & the $\pm1$ result, spread back over the moves by the critic \\
\bottomrule
\end{tabular}
\caption{The learning signal in each stage. Every reward comes from the
end-of-game result; only distillation learns from something else.}
\label{tab:signal}
\end{table}

\subsection{Phase 1: the specialist teachers}\label{sec:phase1}
Before any league training, five algorithms are trained separately for
250{,}000 episodes each, all against the uniform-random opponent. All
recipes below use these same checkpoints, and Phase 1 is never retrained, so
differences between recipes come only from which checkpoints each recipe uses
and how its league picks opponents. The five algorithms optimise different
objectives (Table~\ref{tab:phase1}), so their losses cannot be compared with
each other, and we go through them one at a time. Two of our recipes change the
teacher set, which is why these details matter.

\begin{table}[h]\centering
\begin{tabular}{llll r}
\toprule
Algorithm & What it optimises & Logged losses & Changes during training & Win\% \\
\midrule
PPO   & clipped surrogate + critic & \texttt{policy}, \texttt{value}        & learning rate (fixed schedule) & \textbf{84.1} \\
GRPO  & group-relative, no critic  & \texttt{policy}, \texttt{kl}           & KL coeff.\ $\beta$ (rule-based) & 82.4 \\
SAC   & max-entropy $Q$ learning   & \texttt{actor}, \texttt{q1}, \texttt{q2} & temperature $\alpha$ (learned) & 77.8 \\
DREAM & regret matching (CFR)      & \texttt{adv}, \texttt{strat}           & --                 & 73.6 \\
DPO   & pairwise preference        & \texttt{dpo}                           & --                 & 63.0 \\
\bottomrule
\end{tabular}

\caption{The five specialists. ``Win\%'' is the win rate against uniform
random play, averaged over the final 10\% of training. ``Changes during
training'' names the setting that changes over the run and how: PPO's
learning rate follows a fixed schedule, GRPO's KL coefficient is adjusted by a
rule, and SAC's temperature is learned by gradient descent
(Figure~\ref{fig:phase1coef}, right).}
\label{tab:phase1}
\end{table}

\begin{figure}[h]\centering
\includegraphics[width=0.74\textwidth]{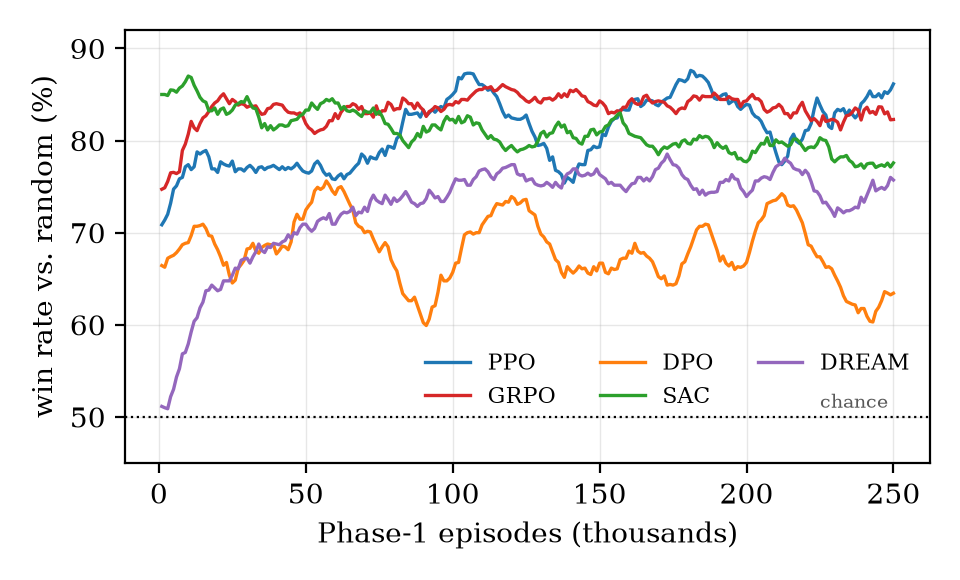}
\caption{Phase-1 specialists, win rate against uniform random play. The five
learn different policies, but this ordering is not the same as their ordering
as teachers (Section~\ref{sec:distill}).}
\label{fig:phase1win}
\end{figure}

\begin{figure}[t]\centering
\includegraphics[width=\textwidth]{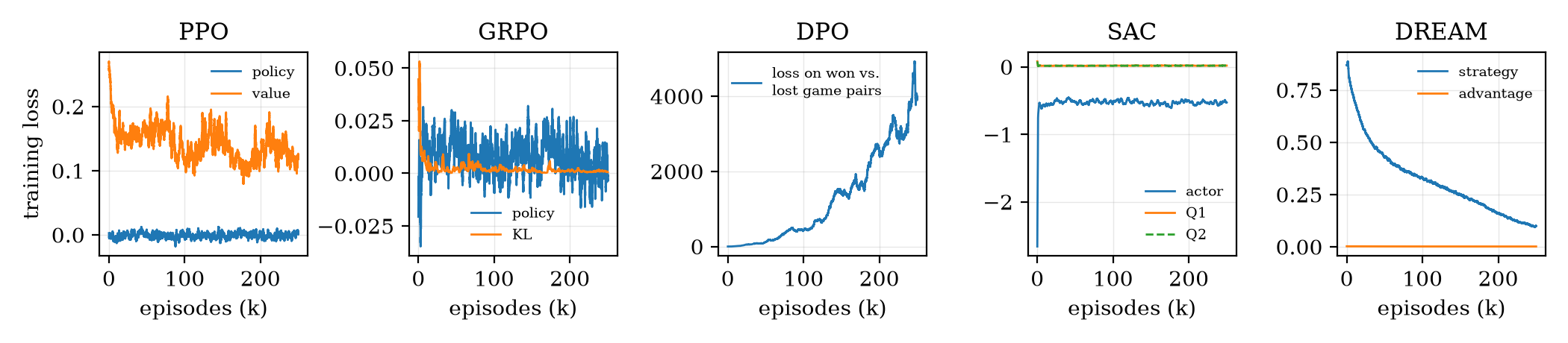}
\caption{Phase-1 training losses, each algorithm on its own axes because a
DPO preference loss, an SAC $Q$-loss and a PPO clipped surrogate are on
different scales. The DPO panel is the preference loss on won-versus-lost game
pairs, and the SAC panel shows both twin critic losses ($Q_1$, $Q_2$), which
almost overlap. The $y$ ranges differ by up to five orders of magnitude
between panels.}
\label{fig:phase1loss}
\end{figure}

\begin{figure}[t]\centering
\includegraphics[width=\textwidth]{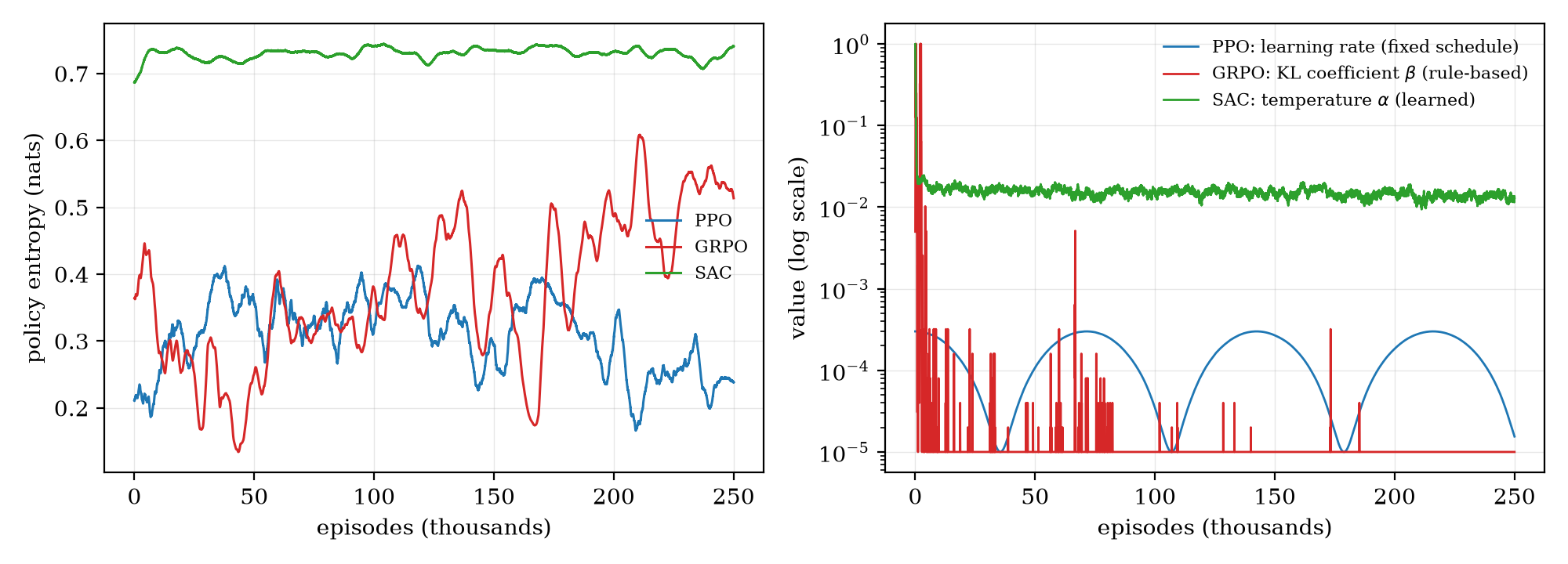}
\caption{\emph{Left}: policy entropy for the three specialists that log it.
SAC is flat by construction because its temperature is tuned to hold a target
entropy; GRPO rises and PPO ends near where it began. \emph{Right}: the
settings that change during training, on a log scale. PPO's learning rate
follows a fixed cosine schedule that cycles about three and a half times (no
restarts); GRPO's KL coefficient collapses five orders of magnitude, from
its $1.0$ cap to its floor, where it sits for $98\%$ of updates after
$6{,}000$ episodes, apart from brief spikes; SAC's learned temperature settles
almost immediately.}
\label{fig:phase1coef}
\end{figure}

\paragraph{PPO:} PPO has two parts: a policy that picks moves, and a critic
that estimates the chance of winning from the current position. The policy is
trained with Eq.~\ref{eq:ppo-loss}; ``clipped'' means
each update can only change the policy a little. Unlike GRPO, each move gets
its own advantage from the critic: whether that move raised or lowered the
critic's estimate of the chance of winning. So PPO can push down a bad move
even inside a game it won, which GRPO cannot. PPO has no KL penalty; its
critic is trained with a separate value loss. Its policy loss stays
near zero (final-10\% mean $6.5\times10^{-6}$). This is expected because of how
PPO prepares its advantages. A move's advantage says how much better it turned out than the critic expected. Before
each update PPO rescales the advantages in the batch so they average zero and
have a spread of one. This way, moves that did better than the batch average
are made more likely and moves that did worse are made less likely, even when
the critic's estimates are still off, and the update step stays about the same
size through training. Because the advantages average zero, the loss also
averages out near zero while the policy is still improving
(Section~\ref{sec:curves} explains the same effect for GRPO). The critic's
value loss, which measures how far its win estimates are from the real
results, settles at $0.126$; Section~\ref{sec:curves} explains why it cannot go
much lower. PPO is also the only specialist
whose learning rate (the size of each update step) follows a timetable fixed
before training: a cosine schedule that falls from $3\times10^{-4}$ to
$10^{-5}$ over $12{,}500$ updates, rises back over the next $12{,}500$, and
repeats. Over the run's $86{,}404$ updates it goes through about three and a
half of these cycles (Figure~\ref{fig:phase1coef}, right).

\paragraph{GRPO:} GRPO has no critic. Instead, the agent plays a group of
$16$ games against the same opponent, and each game's result is compared with
the group average (Eq.~\ref{eq:grpo-adv}). For example, if it wins $10$ of $16$ games, each won game gets an advantage of $+0.78$ and each lost game $-1.29$.
Every move in a game gets that game's advantage, so GRPO cannot tell which move
won or lost the game. Because the average is taken against one opponent, a win
against a weak opponent counts for little and a loss counts for a lot, and the
reverse holds against a strong opponent. This is why no critic is needed. The
policy is updated with the same clipped loss as PPO, plus a KL penalty
(Eq.~\ref{eq:grpo-loss}). KL measures how far the policy has moved from a
frozen reference copy, and the coefficient $\beta$ sets how strongly that
movement is penalised. $\beta$ is adjusted by a
fixed rule after every update: it is doubled when the KL to the reference is
above $1.5\times$ its $0.01$ target, halved when it is below the target divided
by $1.5$, and kept between $10^{-5}$ and $1.0$. It starts at $0.01$, is doubled
to its $1.0$ cap within about $150$ episodes, and is then halved repeatedly
down to its floor of $10^{-5}$; after the first $6{,}000$ episodes it sits at the floor for $98\%$
of updates. After that the KL penalty does almost nothing: the measured KL is
above its $0.01$ target only in roughly the first $12{,}000$ episodes and ends
about ten times below it (final-10\% mean $0.0009$), so GRPO trained almost
unconstrained. Its entropy rises over training, unevenly,
from about $0.32$ nats in the first $25{,}000$ episodes to $0.53$, while PPO's
ends roughly where it began ($0.26$ to $0.25$, peaking near $0.36$ mid-run).

\paragraph{SAC:} Twin $Q$ critics and a maximum-entropy actor. Two things in
its panel need explaining. First, the actor loss is negative
($-0.53$). The loss is
$\sum_a \pi_\theta(a|s)\,\big(\alpha \log \pi_\theta(a|s) - Q(s,a)\big) = -\alpha H - \mathbb{E}_\pi[Q]$.
Here $H$ is the entropy, which
measures how spread out the move choice is and is never negative, and
$\mathbb{E}_\pi[Q]$ is the critic's average value of the moves the policy picks
(using the smaller of the two critics). Since $-\alpha H \le 0$, the loss can
only be positive if the critic expects the agent's moves to lose, by more than
$\alpha H$. An untrained critic outputs values near zero, so the loss is
already negative at the start: it is $-1.61$ at the first update and never
rises above $-0.16$ in the whole run. At the end of training the entropy part
is tiny ($\alpha H = 0.0136 \times 0.724 \approx 0.01$), so $-0.53$ mostly
means the critic values the agent's chosen moves at about $+0.52$, where a win
is $+1$ and a loss is $-1$. Second, the two $Q$-losses track
each other closely down to $0.018$, which means the twin critics agree. SAC
learns its temperature $\alpha$ by gradient descent, with its own optimiser,
going from an initial $1.0$ down to $0.014$. The target entropy is half the
maximum possible entropy over the legal moves: $\alpha$ goes up when entropy is
below this target and down when it is above. Because $\alpha$ holds entropy at
the target, SAC is the one algorithm with a flat entropy curve.

\paragraph{DREAM:} A Deep-CFR-style regret minimiser, and the only specialist
that trains two networks for different jobs. The \emph{advantage} network
fits per-iteration regrets and its loss stays flat and tiny ($\approx0.0014$)
because its target is rebuilt every iteration, so it is always fitting a fresh
small problem. The \emph{strategy} network accumulates the average policy over
all iterations, and its loss decays steadily from about $0.9$ to $0.11$ because
that target settles as the average fills in.

\paragraph{DPO:} DPO learns from pairs of the agent's own games: a game it
won gives the preferred trajectory and a game it lost gives the dispreferred
one. For each trajectory it sums $\log \pi_\theta - \log \pi_{\text{ref}}$ over
the agent's moves, where $\pi_{\text{ref}}$ is a frozen copy refreshed every
500 updates. The implicit reward margin is
$m = \beta\,(\text{won} - \text{lost})$ with $\beta = 0.1$, and the loss is $-\log \sigma(m)$: close to
zero when the policy favours the won game's moves, and roughly $|m|$ when it
favours the lost game's moves. The loss grows from below $1$ at the start to a
final-10\% mean of $3{,}503$, with a peak near $12{,}600$, so the margin
becomes large and negative: the policy ends up favouring the moves from lost
games. DPO only works if the same move is always preferred, and here it is
not. In a game this luck-heavy the same action in the same state often
appears in both won and lost games: playing a Skip with four cards left is a
good move whether or not the deck kills you later. The objective then pushes
identical states in opposite directions and the logits diverge. DPO is also the weakest specialist at $63.0\%$. We think the
cause is how the preference pairs are built from a random end-of-game reward. Recipe \texttt{A} uses DPO as one of its teachers, which may
be part of why the diverse teacher set did worse in
Section~\ref{sec:distill}.

\paragraph{Entropy ordering across phases:} Entropy measures how spread out
an agent's move choice is: low entropy means it almost always picks the same
move, high entropy means it still tries several. In Phase~1, GRPO ends with
more spread than PPO ($0.53$ nats against $0.25$). In Phase~2 it is the other
way round: the PPO-league runs end at $0.566$--$0.630$ and the GRPO-league
runs at $0.165$--$0.276$ (Section~\ref{sec:curves}). So the same two
optimisers, on the same game, swap places between the two phases. The entropy
bonus cannot explain this, because it is the same $0.01$ for both optimisers
in both phases. Two other things do change between the phases: the opponents,
and the network (a 256-wide single head in Phase~1, the 512-wide mixture of
experts in Phase~2).

The entropy floor (Section~\ref{sec:update}), which pushes the agent harder
to try other moves when it starts picking the same move almost every time,
also matters here. The GRPO-league runs hit this floor often. In
\texttt{B-grpo}, the run that produced the champion, the rule was on for
$20.9\%$ of updates, and in \texttt{D} for $15.8\%$ ($1.8$--$20.9\%$ across
the five GRPO-league runs). The PPO-league runs almost never hit it. So
without this rule, GRPO's entropy in Phase~2 would probably have dropped even
lower. Because of this, we do not conclude that one optimiser keeps more
spread than the other in general. The same drop in entropy under GRPO has
also been seen when training language models~\citep{yu2025dapo,cui2025entropy}.

\subsection{How an agent is built}\label{sec:build}
Each agent is a mixture of experts: a shared backbone $f_\theta(s)$ feeds $E$
expert heads, and a sparse top-2 gate picks which two to use. We write
$\pi_\theta(a \mid s)$ for the resulting action distribution over the legal
actions.

\paragraph{Stage 1, distillation:} Expert $i$ is trained to copy teacher $i$
directly. For a neural teacher we read its logits $z_i(s)$ out of its network;
for the rule-based bot, which has no logits, we sample its action
distribution $\hat p_i(s)$ empirically. With temperature $T$ and expert
logits $e_i(s)$:
\begin{equation}
\mathcal{L}_{\text{distill}}
= \underbrace{\sum_{i \in \mathcal{N}} \mathrm{KL}\!\left(
    \sigma\!\left(\tfrac{z_i}{T}\right) \,\big\|\,
    \sigma\!\left(\tfrac{e_i}{T}\right)\right)}_{\text{neural teachers}}
+ \underbrace{\sum_{i \in \mathcal{H}} \mathrm{KL}\!\left(
    \hat p_i \,\big\|\, \sigma(e_i)\right)}_{\text{heuristic, } T=1}
+ \lambda_{\text{feat}} \sum_{i \in \mathcal{N}}
    \big\| g_i(f_\theta(s)) - f_i(s) \big\|_2^2 ,
\label{eq:distill}
\end{equation}
where $g_i$ is a projection head that lets a 512-wide student match a
256-wide teacher's features $f_i(s)$. Each expert learns from its own teacher
at the same time, and the pairing never changes. The gate is not trained here;
it learns which experts to use later, during league play.

Here is where each part of Eq.~\ref{eq:distill} comes from and how it is
computed, using small made-up numbers for one position with three legal moves.

\textbf{KL:} All three parts compare the expert with its teacher. The first
two use the KL divergence, $\mathrm{KL}(p \,\|\, q) = \sum_a p(a) \ln
\big(p(a)/q(a)\big)$, where $p$ is the teacher's move probabilities and $q$ is
the expert's. It is $0$ when the two match and grows as they differ. Each move
is weighted by the teacher's probability, so mistakes on moves the teacher
likes cost the most, and moves the teacher never plays (including illegal
moves) add nothing. Copying a policy by minimising this KL is standard policy
distillation~\citep{rusu2015policy}.

\textbf{Neural teachers (first term):} For PPO, GRPO, DPO, SAC and DREAM we
read the logits $z_i$ (the raw score for each move before it is turned into a
probability) straight from the teacher's network. Both the teacher's and the
expert's logits are divided by a temperature $T = 2$ before the softmax
$\sigma$. This idea comes from \citet{hinton2015distilling}: dividing by $T$
makes the probabilities flatter, so the expert also learns how the teacher
ranks its second and third choices, not only its favourite move. For example,
teacher logits $[4, 2, 0]$ give probabilities $[0.87, 0.12, 0.02]$ at $T=1$
but $[0.67, 0.24, 0.09]$ at $T=2$. The order stays the same, but the backup
moves now carry enough weight to matter in the loss. If the expert's logits
are $[3, 2, 1]$, its probabilities at $T=2$ are $[0.51, 0.31, 0.19]$ and
\[
\mathrm{KL} = 0.67\ln\tfrac{0.67}{0.51} + 0.24\ln\tfrac{0.24}{0.31}
+ 0.09\ln\tfrac{0.09}{0.19} \approx 0.18 - 0.06 - 0.07 = 0.06 .
\]
This is computed once for each of the five neural teachers, each with its own
expert. Hinton et al.\ also multiply this term by $T^2$ to undo the smaller
gradients that come with a softer target; our code does not, so these five
terms push the student about $4\times$ less strongly than the next one.

\textbf{Rule-based bot (second term):} The rule-based bot is a set of
if-then rules with no network, so it has no logits to read. Instead we ask it
$16$ times in the same position and count its answers. If it plays move 1
twelve times and move 2 four times, $\hat p = [0.75, 0.25, 0]$. No temperature
is used ($T = 1$), because there are no logits to divide. With the same expert
logits, the expert's probabilities at $T=1$ are $[0.67, 0.24, 0.09]$, so
$\mathrm{KL} = 0.75\ln\tfrac{0.75}{0.67} + 0.25\ln\tfrac{0.25}{0.24} + 0
\approx 0.10$. Move 3 adds nothing directly because the bot never played it;
the expert is still pushed off it, because raising move 1 lowers the others.

\textbf{Internal state (third term):} The first two terms teach each expert
what move to play. The third teaches the shared backbone to look at the
position the way each teacher does, an idea that goes back to
FitNets~\citep{romero2015fitnets}. Each neural teacher's network turns the
position into $256$ numbers, its features $f_i(s)$; the student's backbone
makes $512$. A small projection layer $g_i$, one per teacher, maps the
student's $512$ numbers down to $256$ so they can be compared, and the loss is
the mean squared difference. For example, with four numbers instead of $256$,
teacher $[0.5, -1.0, 2.0, 0.0]$ and projected student $[0.7, -0.6, 1.5, 0.1]$
differ by $[-0.2, -0.4, 0.5, -0.1]$, whose squares average $0.115$. This term
is multiplied by $\lambda_{\text{feat}} = 0.1$ so it stays a helper next to
the two KL terms. The rule-based bot has no network and so no features, and is
left out of this term.

\textbf{Putting it together:} For each position the loss adds the five
neural-teacher KLs, the rule-based KL and $0.1$ times the five feature errors.
It is averaged over a batch of positions, and one gradient step then updates
all six experts, the shared backbone and the five projection layers together.

\textbf{Two parts:} Distillation runs in two parts with the same loss. In
part~1 ($2{,}500$ steps), the teachers play $500$ games against each other and
the experts learn on the positions from those games. But the student makes
mistakes the teachers would not, and then ends up in positions the teachers
never reached, so part~1 never showed it what to do there. In part~2
($1{,}250$ steps), the student plays $500$ games itself, and we ask each
teacher what it would have played in the positions the student reached. This
is DAgger~\citep{ross2011dagger}.

\paragraph{Stage 2, league play:} The distilled network, gate included, is
then trained by playing 1.25M games against the league pool of
Section~\ref{sec:pfsp}. Every recipe is run twice from the same Stage-1 output:
once updated with GRPO (Eq.~\ref{eq:grpo-loss}, one $64$-game block per update)
and once with PPO and its value head (Eq.~\ref{eq:ppo-loss}); the PPO branch loads the GRPO branch's
distilled checkpoint, so the two start from byte-identical weights. Games are
collected as $8$ concurrent blocks of $64$, a frozen snapshot of the learner
joins the pool every $10{,}000$ games, and a checkpoint is saved at the same
interval. Both branches use learning rate $3\times10^{-4}$, an entropy bonus of
$0.01$, a load-balance coefficient of $0.01$ on the gate, and the same guards:
an update with a non-finite loss or gradient is skipped, and a GRPO update
that moves the policy too far from the one that collected the block is rolled
back. A third guard is the entropy floor
(Section~\ref{sec:update}; Section~\ref{sec:phase1} reports how often it was
active). No continual-learning penalty is used (EWC $\lambda = 0$). Recipe
\texttt{D} skips Stage~1 entirely: both its GRPO run and its PPO run start
from the same saved file of random weights.

\textbf{The two training modes:} The recipes differ in which teachers and
opponents they use (Table~\ref{tab:recipes}). The mode is only how the
network is updated from the games it plays, and every recipe is trained in
both modes from the same starting network.

\emph{GRPO mode:} The learner plays a block of $64$ games against one
opponent, compares each game's result with the block average
(Eq.~\ref{eq:grpo-adv}), and takes one update step. Eight blocks are played at
the same time with the same weights and then used one after another, so the
1.25M games give $19{,}531$ updates. The later blocks in each round were
played by slightly older weights, and the ratio $\rho$ in
Eq.~\ref{eq:grpo-loss} corrects for that.

\emph{PPO mode:} The learner first plays all eight blocks ($512$ games). Its
critic estimates the chance of winning at each position, so every move gets
its own advantage. It then trains on those $512$ games for four passes in
small batches, about $2{,}440$ rounds of updates over the run. If the policy moves
too far from the one that played the games (approximate KL above $0.15$,
three times the $0.05$ target), the remaining passes are skipped.

\subsection{How opponents are chosen}\label{sec:pfsp}
Matchmaking is blocked prioritised fictitious self-play. One opponent is
chosen and then held for $B = 64$ consecutive games, so the learner faces a
fixed opponent long enough for within-block statistics to be meaningful.

Let $x_o$ be the learner's mean score against opponent $o$ over the last $W$
recorded games. Following AlphaStar~\citep{vinyals2019alphastar} we weight by
$f_{\text{var}}$, which peaks at an even matchup:
\begin{equation}
w(o) = \max\big(x_o (1 - x_o),\; 0.01\big), \qquad
x_o = \frac{1}{W}\sum_{k=1}^{W} r_{o,k},
\label{eq:pfsp}
\end{equation}
with $W = 50$ and $w(o) = 0.25$ for an opponent never yet played (an assumed
coin flip, the maximum of $f_{\text{var}}$). On each matchmaking decision, with
probability $0.7$ the opponent is drawn with probability
$w(o) / \sum_{o'} w(o')$; otherwise, with probability $0.3$, it is drawn
uniformly from the candidate set, ignoring $w$. The aim is to prefer opponents
the learner beats about half the time, since those games are the most
informative, instead of always picking the hardest opponent, which would give
a weak learner games it almost always loses.

\paragraph{Fixed-opponent floor:} The pool mixes fixed opponents
(teachers, the rule-based bot, \texttt{random}) with snapshots of the
learner itself, added every 10{,}000 games and never removed. Snapshots
therefore end up making up most of the pool. Our one new flag acts before
Eq.~\ref{eq:pfsp}: with probability $F$, restrict the candidate set to
non-snapshots,
\begin{equation}
\mathcal{O}_{\text{candidates}} =
\begin{cases}
\{o : \neg\,\textsc{IsSnapshot}(o)\} & \text{with probability } F,\\[2pt]
\mathcal{O}_{\text{all}} & \text{otherwise,}
\end{cases}
\label{eq:minfixed}
\end{equation}
falling back to $\mathcal{O}_{\text{all}}$ if the restricted set is empty.
$F = 0$ for every recipe except \texttt{A-plus}, where $F = 0.40$. This sets a
minimum sampling probability; it does not force the realised fraction to any
value. Section~\ref{sec:exposure} measures the fraction it actually produced.

\subsection{How the policy is updated}\label{sec:update}
\textbf{GRPO:} A block of $G$ games against one opponent forms exactly one group-relative
update. Game $j$ returns $R_j \in \{-1, +1\}$, and every step in a game shares
that game's advantage:
\begin{equation}
A_j = \frac{R_j - \mu_R}{\sigma_R + \varepsilon}, \qquad
\mu_R = \frac{1}{G}\sum_{k} R_k, \quad
\sigma_R = \mathrm{std}(R_{1:G}), \quad \varepsilon = 10^{-8}.
\label{eq:grpo-adv}
\end{equation}
Because the mean is taken within a block, opponent difficulty cancels out: $A_j$ asks ``was this game better than my typical result
against this opponent'', so no value head is needed. The policy is then
updated with a clipped surrogate,
\begin{equation}
\mathcal{L}_{\text{policy}} = -\,\mathbb{E}\Big[
\min\big(\rho A,\; \mathrm{clip}(\rho, 1-\epsilon, 1+\epsilon)\,A\big)\Big]
+ \beta\,\mathrm{KL}\!\left(\pi_\theta \,\|\, \pi_{\text{ref}}\right)
- c\,H(\pi_\theta),
\label{eq:grpo-loss}
\end{equation}
where $\rho = \exp\big(\log \pi_\theta(a|s) - \log \pi_{\text{old}}(a|s)\big)$,
with $\epsilon = 0.2$, a KL leash to a periodically refreshed frozen reference
$\pi_{\text{ref}}$, and the log-ratio clamped to $[-20, 20]$ before
exponentiation so $\rho$ cannot overflow. $H(\pi_\theta) = -\sum_a \pi_\theta(a|s)
\log \pi_\theta(a|s)$ is the entropy of the policy, which is $0$ when it always
picks the same move and grows as it spreads its choice over more moves.
Subtracting $c\,H$ gives a small reward for keeping some spread, with
$c = 0.01$. Our GRPO is not exactly the original. The original
GRPO~\citep{shao2024deepseekmath} has no entropy term and uses a fixed $\beta$.
From PPO~\citep{schulman2017ppo} we take the entropy bonus and the rule that
adjusts $\beta$ based on the KL (Section~\ref{sec:phase1}), and we compute the
KL exactly over all legal moves, which is cheap with $186$ actions. In Phase~2
we also add an entropy floor: whenever the entropy falls below $0.15$, $c$ is
multiplied by $4$ for that update. This is similar to the adaptive entropy
coefficient of Skywork-OR1~\citep{he2025skywork}, which raises the coefficient
when entropy drops below a target.

\textbf{PPO:} The PPO branch uses the same clipped surrogate, with three
differences. First, each move gets its own advantage from the critic instead
of sharing its game's advantage. The critic $V(s)$ estimates the chance of
winning from each position, and the advantage measures whether a move raised
or lowered that estimate:
\begin{equation}
\delta_t = r_t + \gamma V(s_{t+1}) - V(s_t), \qquad
\hat A_t = \delta_t + \gamma \lambda \hat A_{t+1},
\label{eq:ppo-adv}
\end{equation}
with $\gamma = 0.99$ and $\lambda = 0.95$. This is generalised advantage
estimation (GAE)~\citep{schulman2016gae}. The advantages of all the games in
an update are then rescaled to average zero with a spread of one. Second,
there is no KL penalty; instead the critic is trained alongside the policy
with a value loss:
\begin{equation}
\mathcal{L}_{\text{PPO}} = -\,\mathbb{E}\Big[
\min\big(\rho \hat A_t,\; \mathrm{clip}(\rho, 1-\epsilon, 1+\epsilon)\,\hat A_t\big)\Big]
+ 0.5\,\big(V(s_t) - \hat R_t\big)^2
- c\,H(\pi_\theta),
\label{eq:ppo-loss}
\end{equation}
where $\hat R_t = \hat A_t + V(s_t)$ is the critic's target, computed before
the advantages are rescaled. Third, PPO trains on the same games for four
passes in batches of $64$ moves, instead of one step per block;
Section~\ref{sec:build} describes how this runs in Phase~2.

\section{Training Curves}\label{sec:curves}
This section goes through what each loss measures and what it shows. Several of
the curves can look wrong at first, so we explain why they behave as they do.

\begin{figure}[t]\centering
\includegraphics[width=0.92\textwidth]{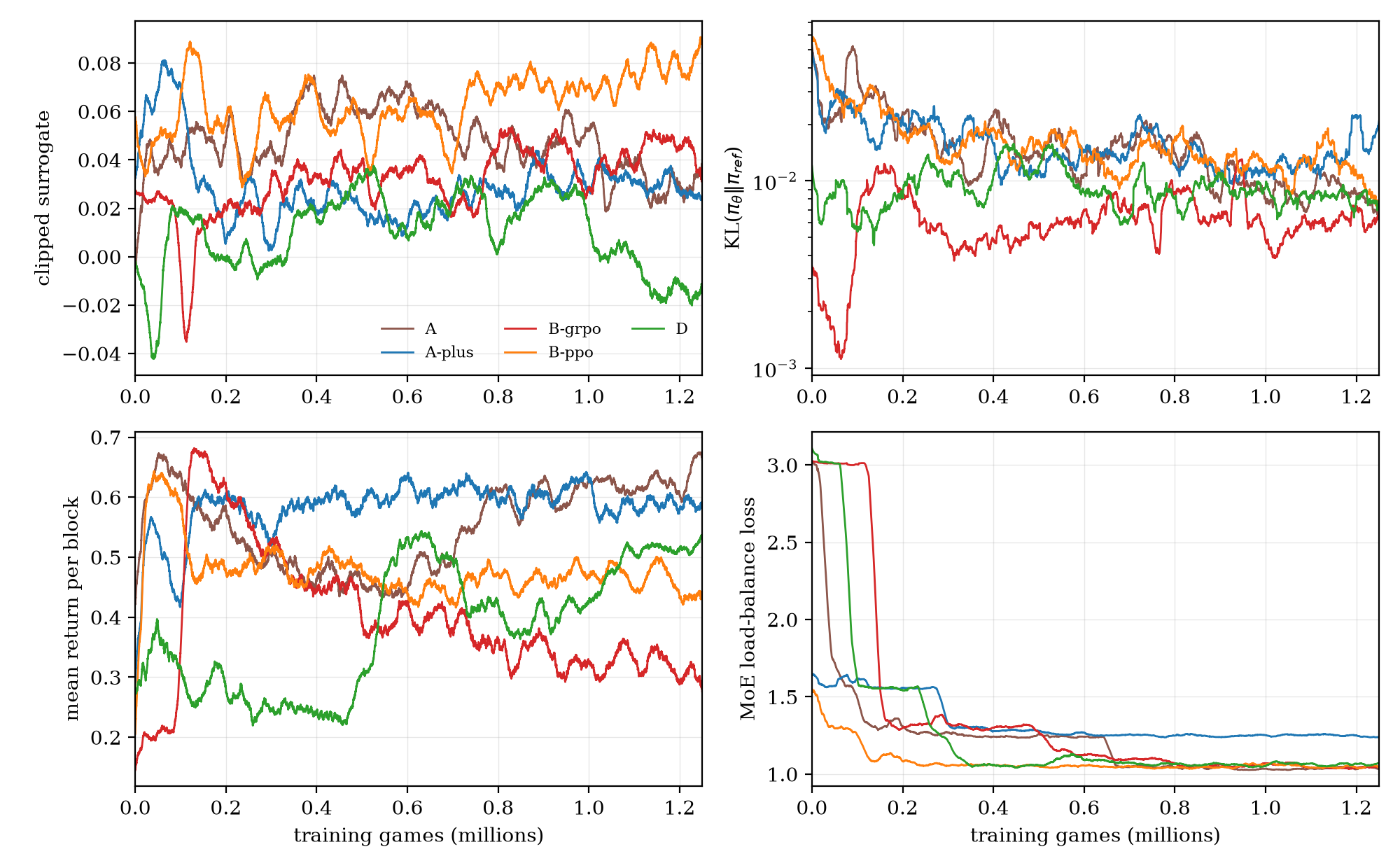}
\caption{Training curves for all five GRPO-league runs, smoothed.
Only \texttt{policy\_loss} and \texttt{lb\_loss} here are shared with the
PPO-league branches (Figure~\ref{fig:trainppo}); \texttt{kl} and
\texttt{group\_return\_mean} exist only under GRPO, because GRPO has no value
head and PPO has no frozen reference policy, so the two sets of runs are
shown in separate figures. The clipped
surrogate (Eq.~\ref{eq:grpo-loss}) stays within $\pm0.09$ of zero throughout,
two orders of magnitude below a supervised loss, because the group-relative
advantage of Eq.~\ref{eq:grpo-adv} is centred within each block. The KL to the
frozen reference spans $0.006$ (\texttt{B-grpo}) to $0.016$ (\texttt{A-plus}),
all well below the $0.05$ target every Phase-2 run used. Below
$\text{target}/1.5$ the adaptive controller halves the KL coefficient on each
update, so in all five runs the KL penalty drifts toward its $10^{-5}$ floor
and does little to hold the policy back, the same as in Phase-1 GRPO
(Section~\ref{sec:phase1}). Mean return per block (bottom left) is the learner's
average $\pm1$ outcome against whichever opponent PFSP scheduled: averaged
over the final 10\% of training \texttt{A} reaches $0.632$ (an 81.6\% win
rate) and \texttt{A-plus} $0.591$ (79.5\%) against \texttt{B-grpo}'s $0.319$
(66.0\%). Playing more fixed opponents does not fully explain this ordering:
\texttt{A-plus} plays the most fixed opponents and wins more than
\texttt{B-grpo}, but \texttt{A} wins more still, and \texttt{D}, which plays
almost only its own snapshots, wins $75.8\%$.}
\label{fig:training}
\end{figure}

\paragraph{Clipped surrogate:} In Figure~\ref{fig:training} (top left) it sits
between $+0.03$ and $+0.08$ for four of the five runs, two orders of magnitude
below a typical supervised loss. Two effects explain this.

It is small because Eq.~\ref{eq:grpo-adv} centres the advantage within each
block: half the games in a block beat the block mean and
half fall short, so $\mathbb{E}[A] = 0$ by construction and, for the first
block of each round, where $\rho = 1$, Eq.~\ref{eq:grpo-loss} evaluates to
exactly zero. So the learner learns
whether a game went better than usual against that opponent, not simply
whether it won, and against a fixed opponent that signal averages to zero.

It is positive because the eight blocks in each round are all played with
the same weights but used for updates one after another. By the time the fifth
block is used, the agent has already been updated four times, so it is
learning from games played by a slightly older version of itself, and $\rho$
is no longer $1$. These small differences are random: they are not related to
whether this block's games were won or lost, so on their own they would
average out to zero. But for each move, Eq.~\ref{eq:grpo-loss} keeps the
smaller of two scores: the actual one and one where the change is capped at
$20\%$. So big changes in a won game get cut, while big changes in a lost game
do not. The random differences therefore add up to a score slightly below
zero, and since the loss is the score with its sign flipped, the loss ends up
slightly above zero. Its size therefore
shows how often the clip is active: \texttt{B-ppo} is highest at $0.081$, so
its updates are the ones the clip limits most. Recipe \texttt{D} is the only
run below zero, at $-0.015$, and it is also the only recipe that starts from
random weights. We did not run the experiment that would link the two, so we
report this only as an observation.

\begin{figure}[t]\centering
\includegraphics[width=0.92\textwidth]{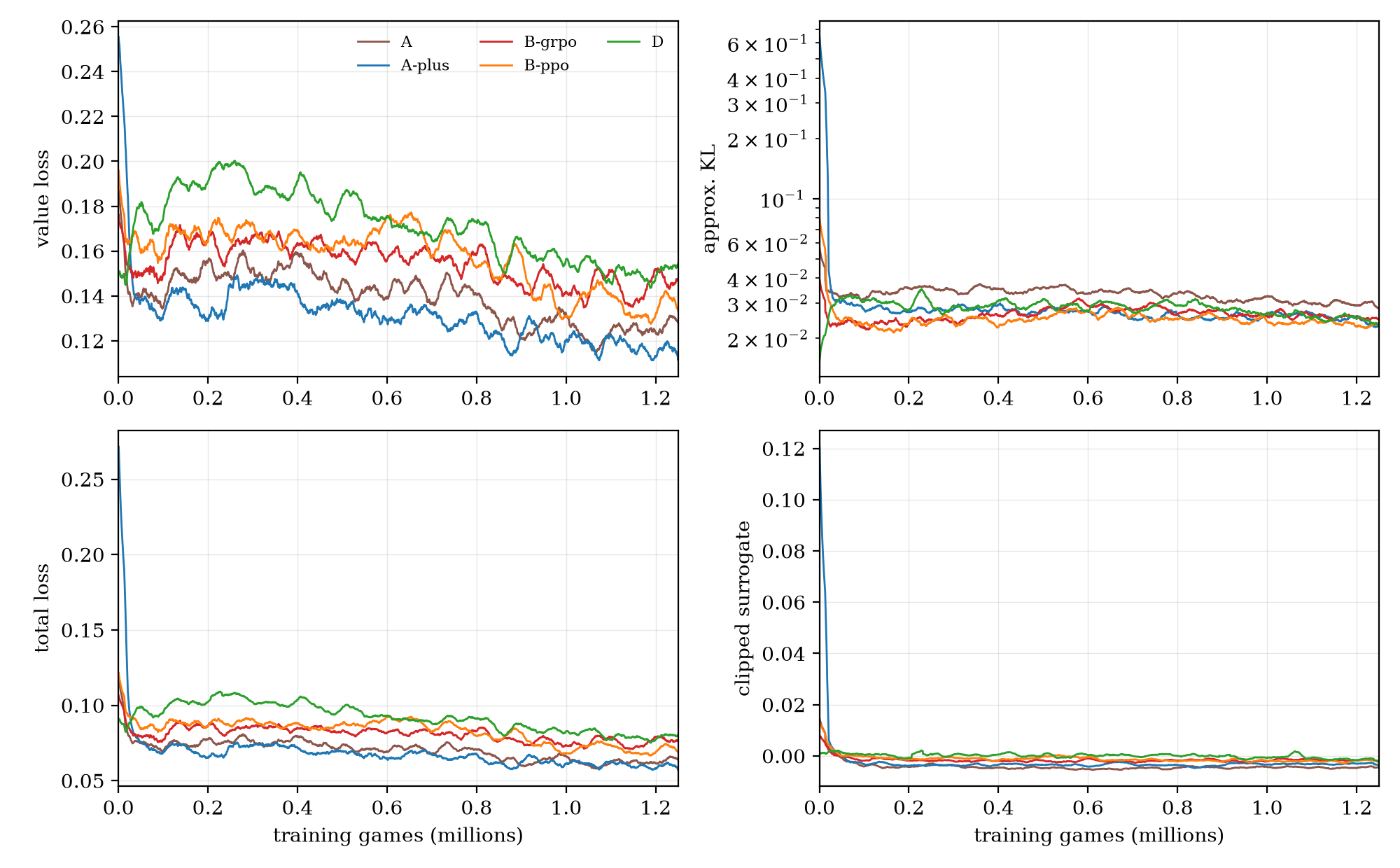}
\caption{The other five runs: the PPO-league runs, with the same five recipes
and colours as Figure~\ref{fig:training}. These log a value loss and an
\texttt{approx\_kl} measured within each update, where the GRPO runs log a
group return and a KL to a frozen reference; only \texttt{policy\_loss} and
\texttt{lb\_loss} appear in both figures. Averaged over the final 10\% of
training, \texttt{A-plus} has the lowest value of both: value loss $0.116$ against \texttt{A} $0.126$, \texttt{B-ppo}
$0.134$, \texttt{B-grpo} $0.143$ and \texttt{D} $0.150$, and the lowest total
loss ($0.060$, against $0.063$--$0.079$ for the rest). \texttt{approx\_kl}
converges to $0.024$--$0.030$ for all five, about half the $0.05$ target and
far below the early-stopping threshold ($3\times$ target, $0.15$) and the
snap-back guard ($10\times$, $0.5$), so on a typical update neither fires and
the clipped ratio alone limits the step. The clipped surrogate (bottom right)
settles slightly below zero here, against ${+}0.03$ to ${+}0.08$ under GRPO.
PPO's advantages are GAE estimates~\citep{schulman2016gae} ($\lambda = 0.95$,
four inner epochs), which carry their own baseline error instead of being
exactly centred within a block, and GAE is known to suffer from high variance in imperfect-information
self-play~\citep{gaeshort2026}. As a result the
pessimistic branch of the $\min$ is picked less consistently.}
\label{fig:trainppo}
\end{figure}

\paragraph{Value loss:} In the PPO runs (Figure~\ref{fig:trainppo}, top left)
it settles at $0.116$--$0.150$ and does not go lower. The value head predicts
the final $\pm1$ outcome from the current observation, and in this game much of
that outcome is still undecided: one Exploding Kitten sits in a 35-card deck,
and drawing it without a Defuse ends the game at once. A strong position can
still lose on the next draw, so the remaining error comes from the game's
randomness. Phase-1 PPO, trained only against the random opponent, ends at a
similar $0.126$, which also points to the game rather than the opponents or
the optimiser.

The two ends of the ordering match fixed-opponent exposure
(Section~\ref{sec:exposure}): \texttt{A-plus}, which plays the most fixed opponents
(final-window share $0.600$ in its PPO branch), has the lowest value loss at
$0.116$, and \texttt{D}, which mostly stops playing its one fixed opponent within
roughly the first 3\% of training (share $0.029$), has the highest at $0.150$.
A fixed opponent is easier to predict than a snapshot of a policy that was
still changing. The middle of the ordering does not follow this: \texttt{B-ppo} and
\texttt{B-grpo} end with almost the same share ($0.307$ and $0.303$) but
different value losses ($0.134$ and $0.143$), and \texttt{A} ($0.126$) has no
exposure measurement at all, so we read exposure as one factor rather than
the whole explanation.

\paragraph{Training win rates:} Eq.~\ref{eq:pfsp} peaks at $x_o = 0.5$, so
PFSP should steer toward even matchups. It does not get there: final-10\% mean returns in the GRPO
league are $0.632$ (\texttt{A}), $0.591$ (\texttt{A-plus}), $0.516$
(\texttt{D}), $0.460$ (\texttt{B-ppo}) and $0.319$ (\texttt{B-grpo}), i.e.\ win
rates of $81.6\%$ down to $66.0\%$. PFSP can only choose even matchups
when such opponents exist. Once the learner is much stronger than
\texttt{random}, \texttt{rule\_based} and the early teacher checkpoints, no
sampling rule can turn those into even games, and $w(o) \geq 0.01$ means they
keep being picked. Weak fixed opponents are only part of the reason, though,
because the returns do not follow the pools: \texttt{D}, whose pool is almost entirely its
own snapshots (fixed share $0.034$), has the second-highest return. Snapshots
are older, weaker copies of the learner, so they are beatable too.

The seat also plays a part. During training the learner
always moves first, and the first mover wins $63.4\%$ of games between the
agents we evaluated (all 612{,}500 super-Elo games, where seats alternate). Against an exact copy of itself
the learner would therefore expect a mean return near $0.27$ rather than $0$.
B-grpo's $0.319$ is only a little above that, so its training games are closer
to even than the raw number suggests, and the other recipes' margins over
their pools are smaller than their returns imply.

\paragraph{Load-balance loss:} $L = E \sum_i \bar p_i^2$ over $E = 6$ experts ranges from $1.0$ when the gate
spreads traffic uniformly to $6.0$ when it collapses onto one expert, so
$E / L$ is an effective expert count. Every run starts more concentrated
than it ends ($L$ between $1.7$ and $3.2$ at the first logged update) and
spreads out. In the GRPO league the final values are $1.04$ for \texttt{A}
and \texttt{B-grpo} ($5.8$ of 6 experts), $1.05$--$1.06$ for \texttt{B-ppo}
and \texttt{D}, and $1.25$ for \texttt{A-plus} ($4.8$ experts). In the PPO
league all five, \texttt{A-plus} included, end at $1.07$, so the
\texttt{A-plus} difference below is a GRPO-league result only.

In the GRPO league, a value close to $1.04$ is not necessarily better.
\texttt{B-grpo}'s six experts are distilled from six checkpoints of the same
algorithm, so they are almost interchangeable and the gate has little reason
to prefer any of them; near-uniform routing is what we would expect from
redundant experts. \texttt{A-plus}, whose experts come from five different
algorithms plus a heuristic, uses fewer of them. We did not run the experiment
that would separate redundant experts from gate collapse, so this is an
observation only.

\begin{figure}[t]\centering
\includegraphics[width=0.49\textwidth]{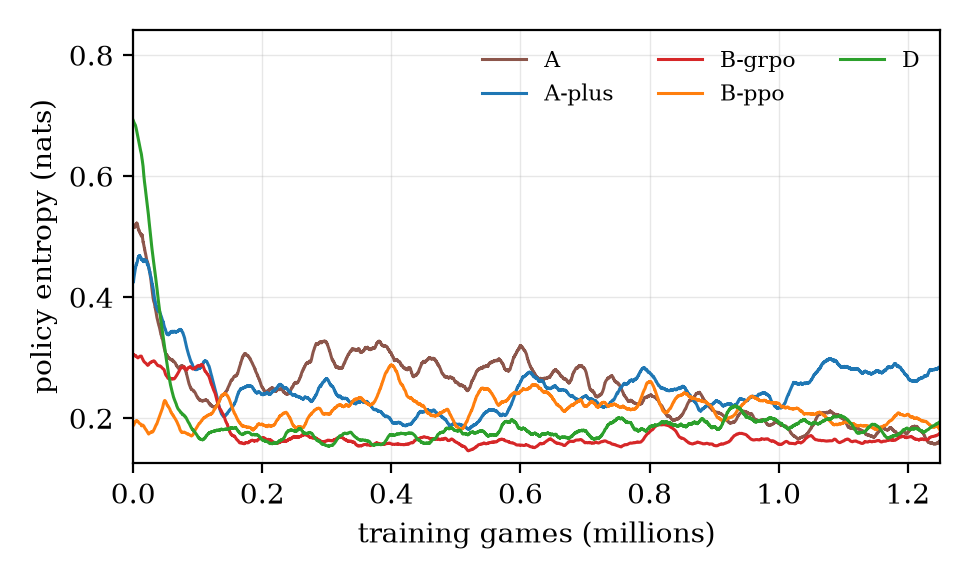}
\hfill
\includegraphics[width=0.49\textwidth]{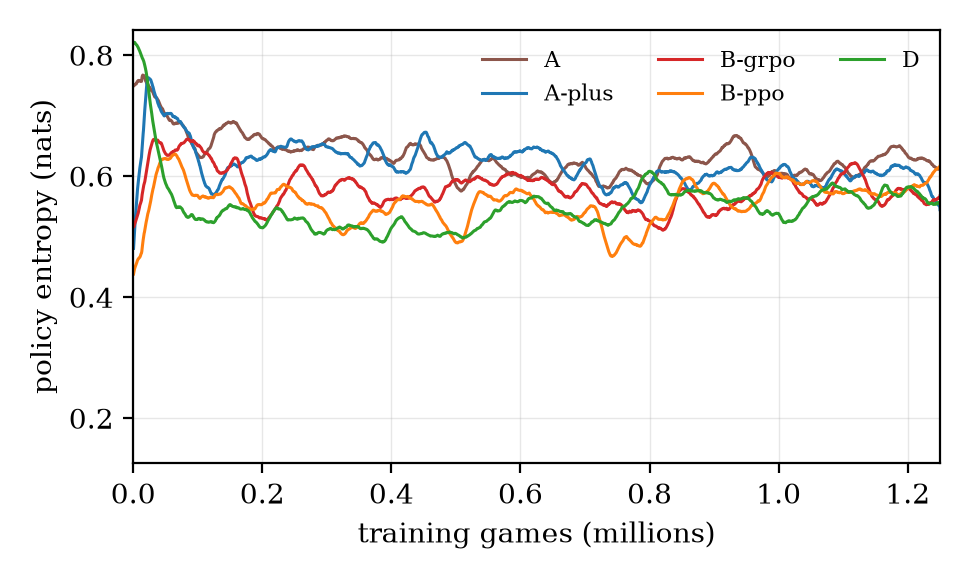}
\caption{Policy entropy: GRPO league (left), PPO league (right), on a shared
$y$ axis. The league algorithm matters more than the recipe. PPO-league
runs end at $0.566$--$0.630$ nats and GRPO-league runs at
$0.165$--$0.276$, a $3.0\times$ gap, and the between-branch difference
($0.391$) is $3.5\times$ the widest within-branch spread ($0.110$). Within the
GRPO panel the clearest comparison is \texttt{A} against \texttt{A-plus},
which differ in exactly one flag: $0.175$ against $0.276$ at an identical
budget. Across all ten runs, entropy does not rise or fall steadily with the
score against outside opponents, so we report only this one pairwise
comparison.}
\label{fig:entropy}
\end{figure}

\paragraph{Entropy:} In Figure~\ref{fig:entropy} the curves split almost
perfectly by league algorithm (GRPO or PPO), not by recipe. This is the
largest effect in the training logs, and it comes from the two optimisers.

\section{Experiments and Results}\label{sec:results}
Section~\ref{sec:protocol} describes how the agents are evaluated and
Section~\ref{sec:ablations} which runs are compared; the rest of this section
reports what we found, one question per subsection. The trained
agents beat both LLMs by a wide margin (Section~\ref{sec:llm}). The recipe that
played fixed opponents most often is the best against outside opponents but
near the bottom against the other trained agents (Section~\ref{sec:ood}), and
Section~\ref{sec:exposure} shows how often each recipe actually faced fixed
opponents. The student distilled from six teachers is weaker than its best
teacher (Section~\ref{sec:distill}), and Section~\ref{sec:moe} looks at what
the mixture of experts does. Ranking by games won and by matches won gives
different winners (Section~\ref{sec:eval}), because some recipes produce
specialists and others generalists (Section~\ref{sec:archetype}).

\subsection{Evaluation protocol}\label{sec:protocol}
All ratings are maximum-likelihood Bradley--Terry~\citep{bradley1952rank}: the
probability that $i$ beats $j$ is
\begin{equation}
\Pr(i \succ j) = \frac{1}{1 + 10^{(r_j - r_i)/400}},
\label{eq:bt}
\end{equation}
and we choose $r$ to maximise the likelihood of every observed game, counting
a draw as half a win, with ratings centred on a mean of 1200 and 95\%
intervals from a 500-sample bootstrap. Seats alternate exactly, so each pair
plays half its games in each seat. Where we report a win rate against a single
fixed baseline we give a Wilson interval and, as a convenience only, the
implied rating gap $\Delta = 400 \log_{10}\!\big(p / (1-p)\big)$, which is just
$p$ expressed on the Elo scale.

\textbf{Match record:} We also rank agents a second way, by head-to-head
matches. In a round robin, all the games between two agents (for example the
$2{,}000$ games of a playoff pair) count as one match, and whoever wins more
than half of those games wins the match. A match split exactly $50/50$ is a
tie. Agents are ranked by the number of matches they win, and if two agents
win the same number, the one with the higher share of games won goes first.
The two rankings answer different questions. Bradley--Terry asks ``who wins
the most games against this field?'', so a big win counts for more than a
narrow one. The match record asks ``who beats the most opponents?'', so a
match won $52\%$ to $48\%$ counts the same as one won $70\%$ to $30\%$. For
example, an agent that wins two matches $70\%$ to $30\%$ and loses two
$45\%$ to $55\%$ wins $57.5\%$ of its games but has a $2$--$2$ record, while
one that wins three matches $52\%$ to $48\%$ and loses one $48\%$ to $52\%$
wins only $51\%$ of its games but has a $3$--$1$ record. Bradley--Terry ranks
the first agent higher; the match record ranks the second one higher.
Section~\ref{sec:eval} shows that this happens in our tournaments.

\textbf{Evaluation stages:} After training, the agents are scored in five
stages. Each stage narrows the field and plays more games per pair, so the
final rankings are more precise. In the first three stages every agent plays
every other agent (a round robin), so $n$ agents give $n(n-1)/2$ pairs. In the
last two, every agent plays only a few fixed opponents.
\begin{itemize}
\item \emph{Qualifiers:} One qualifier per recipe, five in total. Each recipe
has $20$ checkpoints ($10$ saved points along training, for both its GRPO and
its PPO run), and they all play each other: $190$ pairs $\times$ $300$ games
$= 57{,}000$ games per qualifier, or $285{,}000$ for all five. The top $10$
checkpoints of each recipe go on to the super Elo.
\item \emph{Super Elo:} The top $10$ from each of the five recipes, $50$
agents, all play each other in one combined round robin. We call it the super
Elo because it puts the best checkpoints of every recipe into a single rating
list: $1{,}225$ pairs $\times$ $500$ games $=
612{,}500$ games. This stage compares the recipes, and most recipe-level
results in this paper come from it.
\item \emph{Playoff:} The top $20$ of the super Elo play each other again, with
four times as many games per pair: $190$ pairs $\times$ $2{,}000$ games $=
380{,}000$ games. This stage gives a precise ranking of the strongest agents;
the champion and the match records come from it.
\item \emph{Baseline anchor:} The $20$ playoff agents each play two fixed
outside opponents, \texttt{random} and \texttt{rule\_based}: $40$ pairs
$\times$ $1{,}000$ games $= 40{,}000$ games. Because these opponents never
change, the results mean something outside our own field.
\item \emph{LLMs:} Each of the two LLMs plays the $20$ playoff agents plus
\texttt{random} and \texttt{rule\_based}: $22$ pairs $\times$ $200$ games $=
4{,}400$ games per LLM, or $8{,}800$ for both. Fewer games are played here
because every LLM move is slow (Section~\ref{sec:llm}).
\end{itemize}
In total, the evaluation plays $1{,}326{,}300$ games.

\subsection{Ablation design}\label{sec:ablations}
Starting from the Phase-1 specialists of Section~\ref{sec:phase1}, we train
five recipes (Table~\ref{tab:recipes}). Recipe \texttt{A} is the baseline, and
each other recipe changes the teacher set or the league pool. Each recipe is
trained twice from the same starting weights, once in GRPO mode and once in PPO
mode, giving ten runs of 1.25M games each. This includes recipe \texttt{D},
whose two runs both start from the same random weights.

\begin{table}[h]\centering\small\setlength{\tabcolsep}{4pt}
\begin{tabular}{lllll}
\toprule
Recipe & Teachers (distillation step) & League pool & Distinguishing flag & Stage-2 runs \\
\midrule
\texttt{A} & 5 families + rule-based & 30 fixed & (baseline) & GRPO, PPO \\
\texttt{A-plus} & same as \texttt{A} & 30 fixed & \texttt{--min-fixed-fraction 0.40} & GRPO, PPO \\
\texttt{B-grpo} & 6 GRPO stages & 26 fixed, GRPO only & single algorithm & GRPO, PPO \\
\texttt{B-ppo} & 6 PPO stages & 26 fixed, PPO only & single algorithm & GRPO, PPO \\
\texttt{D} & none (random init) & 1 fixed (\texttt{random}) & no distillation & GRPO, PPO \\
\bottomrule
\end{tabular}

\caption{The five recipes. Only the teacher set and the league pool change;
architecture, budget, and all other hyperparameters are held fixed. Every
recipe, including \texttt{D}, is trained twice in Stage~2: once in GRPO mode
and once in PPO mode, both from the same starting weights.}
\label{tab:recipes}
\end{table}

Table~\ref{tab:ablations} lists the comparisons these runs make possible, the
question each one answers, and where the answer is.

\begin{table}[h]\centering\small\setlength{\tabcolsep}{4pt}
\begin{tabular}{p{2.6cm}p{3.6cm}p{5.0cm}p{2.4cm}}
\toprule
Comparison & What changes & Question & Section \\
\midrule
\texttt{A} vs.\ \texttt{A-plus} & fixed-opponent floor, $0 \to 0.40$ & Does playing fixed opponents more often help against outside opponents? & \ref{sec:ood}, \ref{sec:exposure} \\
\texttt{A} vs.\ \texttt{B-grpo} & 6 varied teachers $\to$ 6 GRPO checkpoints, and the pool & Do more varied teachers make a better student? & \ref{sec:distill} \\
\texttt{B-grpo} vs.\ \texttt{B-ppo} & teacher algorithm, GRPO or PPO & Does it matter which algorithm teaches? & \ref{sec:distill} \\
\texttt{D} vs.\ the rest & no distillation (random start) & What does distillation add? & \ref{sec:distill}, \ref{sec:ood} \\
GRPO vs.\ PPO run of each recipe & league update rule & Does the league algorithm matter? & \ref{sec:curves}, \ref{sec:ood} \\
top-2 vs.\ top-1 expert & experts used per move (a probe, not a training run) & Does the mixture of experts change the moves? & \ref{sec:moe} \\
\bottomrule
\end{tabular}
\caption{What each comparison changes, the question it answers, and the
section with the answer.}
\label{tab:ablations}
\end{table}

\textbf{What is not controlled:} \texttt{A} versus \texttt{B-grpo} changes
two things at once: the teachers, and the league pool (26 GRPO checkpoints
instead of 30 opponents from all five algorithms and the rule-based bot), so
this comparison cannot say which of the two made the difference. The other
limits of these comparisons (one seed per recipe, the missing recipe C, and no
exposure log for recipe \texttt{A}) are listed in Section~\ref{sec:limits}.

\subsection{LLM Baseline}\label{sec:llm}
Both LLMs get the agents' exact observation, converted to text with nothing
added or removed, and the same list of legal moves. A reply that cannot be read
or names an illegal move is replaced by a random legal move and counted; it is
never retried. Every reply from both models could be read. Gemma-4-31B named an
illegal move 21 times in 135{,}319 decisions ($0.016\%$), and Qwen3-30B never
did.

\begin{table}[t]\centering
\begin{tabular}{lrr}
\toprule
 & Qwen3-30B-A3B & Gemma-4-31B \\
 & (MoE, 3B active) & (dense, FP8) \\
\midrule
vs.\ 20 trained agents & 10.5\% & 13.3\% \\
\quad ratio & 8.5:1 & 6.5:1 \\
vs.\ \texttt{rule\_based} & 74.0\% & 84.0\% \\
vs.\ \texttt{random} & 50.0\% & 66.0\% \\
fallback rate & 0.0\% & 0.0\% \\
games & 4,400 & 4,400 \\
\bottomrule
\end{tabular}

\caption{Both models are far below the trained agents, which score
85.9--94.4\% against \texttt{random}. Going from a 30B MoE to a 31B dense model
does not close the gap.}
\label{tab:llm}
\end{table}

This does not mean the LLMs play at chance level. Qwen3-30B scores exactly
50.0\% against \texttt{random}, but Gemma-4-31B scores 66.0\%
(Table~\ref{tab:llm}).

\paragraph{Attack stacking:} Stacking means playing an \textsc{Attack} while
already owing turns, so the whole debt passes to the opponent with two turns
added. It is the strongest tempo move in the game. Trained agents use it on
31.9\% of their attacks (230{,}757 of 723{,}852 across all playoff
games); Qwen3-30B uses it on 0.6\% (49 of 8{,}197) and Gemma-4-31B on 3.7\% (246 of
6{,}688), despite a prompt that states the rule verbatim. A player can only
stack when it already owes turns, so the rate also depends on the opponent.
\texttt{random} plays \textsc{Attack} often and so leaves the LLM owing turns
more often: 104 of Gemma's 246 stacks come from its 200 games against
\texttt{random}. Against the 20 trained agents alone, Gemma stacks 2.4\% of
its \textsc{Attack}s and Qwen3-30B 0.4\%. For comparison, \texttt{random}
itself stacks 5.5--6.8\% of its \textsc{Attack}s just by chance (the
\texttt{random} rows in Table~\ref{tab:stacking}). We count an \textsc{Attack} as stacked only when
the player still owes two or more turns at that moment, the condition under
which the engine actually passes on the debt plus two. In its reasoning traces Qwen3-30B discusses deck odds well, but it still does
not apply this rule when it matters (Table~\ref{tab:stacking}).

\begin{table}[h]\centering
\begin{tabular}{lrrl}
\toprule
Player & Stacked / total \textsc{Attack}s & Rate & 95\% CI \\
\midrule
Trained agents (all playoff games) & 230{,}757 / 723{,}852 & \textbf{31.9\%} & $[31.8, 32.0]$ \\
\texttt{random} (vs.\ Qwen3-30B) & 24 / 352 & 6.8\% & $[4.6, 9.9]$ \\
Qwen3-30B-A3B & 49 / 8{,}197 & \textbf{0.6\%} & $[0.5, 0.8]$ \\
\texttt{random} (vs.\ Gemma-4-31B) & 17 / 309 & 5.5\% & $[3.5, 8.6]$ \\
Gemma-4-31B & 246 / 6{,}688 & \textbf{3.7\%} & $[3.3, 4.2]$ \\
\bottomrule
\end{tabular}

\caption{\textsc{Attack} stacking: an \textsc{Attack} played while still
owing two or more turns, so the whole debt passes to the opponent with two
added. Each row counts that player's own \textsc{Attack}s: the trained agents over
all playoff games, each LLM over all its games (against the 20 trained agents,
\texttt{random} and \texttt{rule\_based}), and \texttt{random} over its games
against that LLM. The \texttt{random} rows show how often stacking happens by
chance, when every move is picked at random. Both LLMs stack far less often
than the trained agents. Qwen3-30B also stacks less often than random play,
and the two intervals do not overlap; Gemma-4-31B's interval overlaps random
play's.}
\label{tab:stacking}
\end{table}

\subsection{Results Against Outside Opponents}\label{sec:ood}
The recipes differ in how often they play fixed opponents instead of snapshots
of themselves (Table~\ref{tab:recipes}). Table~\ref{tab:axes} groups the $20$
playoff agents by recipe and compares them against the four outside opponents
and in the playoff.

\begin{table}[t]\centering
\begin{tabular}{lrrrrr}
\toprule
Recipe & $n$ & vs.\ LLMs $\downarrow$ & vs.\ \texttt{random} & vs.\ \texttt{rule\_based} & Playoff M\# $\downarrow$ \\
\midrule
A-plus & 3 & \textbf{7.3\%} & 93.9\% & 97.3\% & 17.7 \\
B-grpo & 9 & \textbf{11.6\%} & 89.6\% & 93.2\% & 5.4 \\
D & 2 & \textbf{12.4\%} & 90.2\% & 94.8\% & 10.0 \\
B-ppo & 1 & \textbf{13.0\%} & 85.9\% & 91.0\% & 18.0 \\
A & 5 & \textbf{14.8\%} & 89.7\% & 95.4\% & 14.0 \\
\bottomrule
\end{tabular}

\caption{A-plus is first on all three measures against outside opponents and
17.7th of 20 in the playoff. ``vs.\ LLMs'' is the mean score the two LLMs achieved
against that recipe's checkpoints, so lower is better. $n$ is how many of
the $20$ playoff agents come from that recipe. ``Playoff M\#'' is the average
rank of those agents in the playoff, where the $20$ trained agents play each
other and are ranked by matches won ($1$ = best, $20$ = last), so lower is
better. A-plus's three agents finish 16th, 17th and 20th, giving $17.7$. Recipe
A, which has A-plus's pool without the floor, is the worst against LLMs.}
\label{tab:axes}
\end{table}

All four outside opponents (a random player, a hand-written rule-based bot,
and two LLMs from different model families) put A-plus first, while the playoff
puts it near the bottom. Below first place, the orderings differ between
columns.

Two comparisons are made here, based on different evidence. The outcome
comparison in Table~\ref{tab:axes} includes \texttt{A}: its
checkpoints were evaluated against all four outside opponent classes like
every other recipe's, and \texttt{A} and \texttt{A-plus} differ in exactly one
flag, so A-plus beating A on every outside-opponent column is the clearest
comparison in the table. The explanation, how much more often A-plus actually
faced fixed opponents, can only be shown against \texttt{B} and
\texttt{D}, because \texttt{A} was trained before the exposure metric existed
(Limitations). Against those, A-plus's final-window fixed-opponent share
($0.576$--$0.600$) is roughly twice B-grpo's ($0.309$) and many times D's.

\textbf{Why this might happen:} An agent that mostly plays snapshots of itself
learns to beat strong players that think like it, which is exactly who it
meets in the playoff. An agent that keeps playing fixed opponents, which play
in very different ways, learns to handle play it has not seen, which is what
an outside opponent like an LLM looks like. So the floor trades strength
against the league for robustness against everyone else. We did not run the
experiment that would confirm this, such as varying the floor over several
values, so this is an explanation that fits the results, not a tested one.

\subsection{How Often Each Recipe Faced Fixed Opponents}\label{sec:exposure}
\texttt{--min-fixed-fraction} places a floor on the share of games played
against fixed, non-snapshot opponents. Every run starts at exactly $1.0$,
because before the first snapshot the pool holds only fixed opponents, and the
fraction then drops as snapshots are added and PFSP starts picking them.

We logged \texttt{league/fixed\_opponent\_fraction} every 100 games over each
1.25M-game run (12{,}500 points per run; all values here are computed from the full logs).
Table~\ref{tab:fixedfrac} reports the mean within three explicit windows: the
first 10\% of training (games 0--125{,}000), the middle 10\%
(562{,}500--687{,}500) and the final 10\% (1{,}125{,}000--1{,}250{,}000),
together with the first game at which the fraction fell below one half.

\begin{table}[t]\centering
\begin{tabular}{lrrrr}
\toprule
 & \multicolumn{3}{c}{\texttt{league/fixed\_opponent\_fraction}} & First game \\
\cmidrule(lr){2-4}
Run & games 0--125k & 562k--688k & 1{,}125k--1{,}250k & below $0.50$ \\
\midrule
\texttt{A-plus} (grpo) & 0.941 & 0.657 & \textbf{0.576} & \textbf{never} \\
\texttt{A-plus} (ppo)  & 0.945 & 0.678 & \textbf{0.600} & \textbf{never} \\
\texttt{B-grpo} (grpo) & 0.909 & 0.448 & 0.309 & 502{,}299 \\
\texttt{B-grpo} (ppo)  & 0.875 & 0.423 & 0.303 & 427{,}599 \\
\texttt{B-ppo} (grpo)  & 0.870 & 0.397 & 0.281 & 375{,}899 \\
\texttt{B-ppo} (ppo)   & 0.876 & 0.422 & 0.307 & 419{,}899 \\
\texttt{D} (grpo)      & 0.448 & 0.058 & 0.034 & 38{,}499 \\
\texttt{D} (ppo)       & 0.412 & 0.049 & 0.029 & 32{,}799 \\
\bottomrule
\end{tabular}

\caption{Realised fixed-opponent exposure. Windows are means over the stated
game ranges, not over sample indices. Both \texttt{A-plus} branches are the
only runs that never spend a majority of their games against their own
snapshots; \texttt{D} mostly stops playing its single fixed opponent (the
uniform-random bot) within roughly the first 3\% of training.}
\label{tab:fixedfrac}
\end{table}

\begin{figure}[t]\centering
\includegraphics[width=0.78\textwidth]{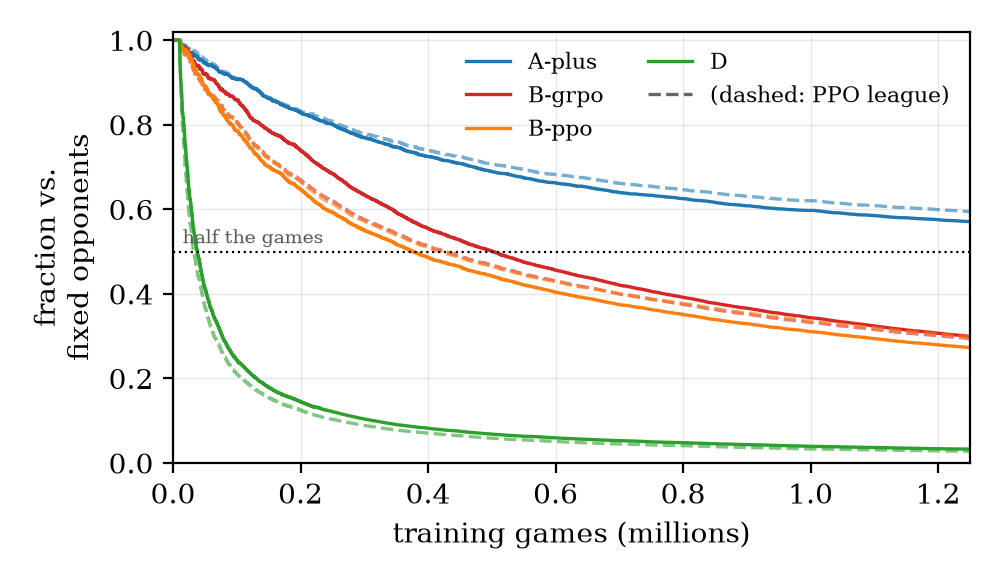}
\caption{Realised fixed-opponent exposure over 1.25M games. All runs start at
$1.0$; \texttt{A-plus} never crosses $0.5$, \texttt{B-grpo} crosses at 40\% of
training (GRPO branch), \texttt{D} at about 3\%. Solid lines are GRPO-league
runs and dashed lines PPO-league runs; recipe \texttt{A} predates the metric.}
\label{fig:fixedfrac}
\end{figure}

Figure~\ref{fig:fixedfrac} shows the full curves. Neither A-plus run ever
falls below $0.50$; both end near $0.58$--$0.60$,
whereas B-grpo crosses one half at 40\% of training in its GRPO branch (34\% in its PPO branch) and D at about 3\%.
A-plus is therefore the only recipe that never spent a majority of its games
against its own snapshots. The floor rarely has to act; its main effect is a
much slower decline.

\subsection{Multi-Teacher Distillation}\label{sec:distill}
The first half of the pipeline assumes that distilling several different
specialists into one mixture gives a better starting point than any single
specialist. We can test this directly, because the distilled student can play
like any other agent, and it and its teachers were rated in the same
tournament.

\paragraph{Distilled student vs. its teachers:} The Stage-1 output
\texttt{super\_distilled}, before any league play, scored Elo $1054$. Its best
teacher, the Phase-1 PPO specialist, scored $1155$, so after copying six
teachers the student was 101 Elo weaker than the best of them. League play
therefore started from below the level of the best teacher.

\paragraph{Diverse vs. single-algorithm teachers:} If diverse teachers helped,
recipe \texttt{A} (five algorithms plus a rule-based bot) should beat recipe
\texttt{B-grpo} (six checkpoints of GRPO alone). It does not: B-grpo is clearly ahead. Averaged over the ten entrants each recipe placed in the 50-agent super
Elo, B-grpo's mean Bradley--Terry rank is $12.6$ against $20.0$ for A. The playoff, ranked by matches won, agrees: B-grpo's nine playoff agents average rank $5.4$ against A's five at $14.0$ (Table~\ref{tab:axes}), and
B-grpo supplies eight of the playoff's top ten (Table~\ref{tab:playoff}).
B-grpo also has no rule-based expert, so on paper it is the less equipped
recipe.

\paragraph{Which algorithm teaches:} On the same super-Elo measure, the widest
gap between recipes is between \texttt{B-grpo} (12.6) and \texttt{B-ppo}
(33.2). Both use six checkpoints of a single algorithm and no rule-based bot;
they differ only in which algorithm supplied the teachers. This gap is larger
than the gap between diverse and single-algorithm teachers, and it goes against
Phase-1 strength: GRPO was the weaker specialist (Elo 1078 against PPO's 1155,
and $82.4\%$ against $84.1\%$ versus random) but made the better teacher.

This matches results outside games: \citet{openmopd2026} report that in LLM post-training a multi-teacher
student, even with each prompt routed to the right teacher, fell short of its
single-domain teachers. Section~\ref{sec:moe} shows a possible reason in our
case: the experts still differ after training, but too little to change many
moves.

\begin{figure}[t]\centering
\includegraphics[width=0.78\textwidth]{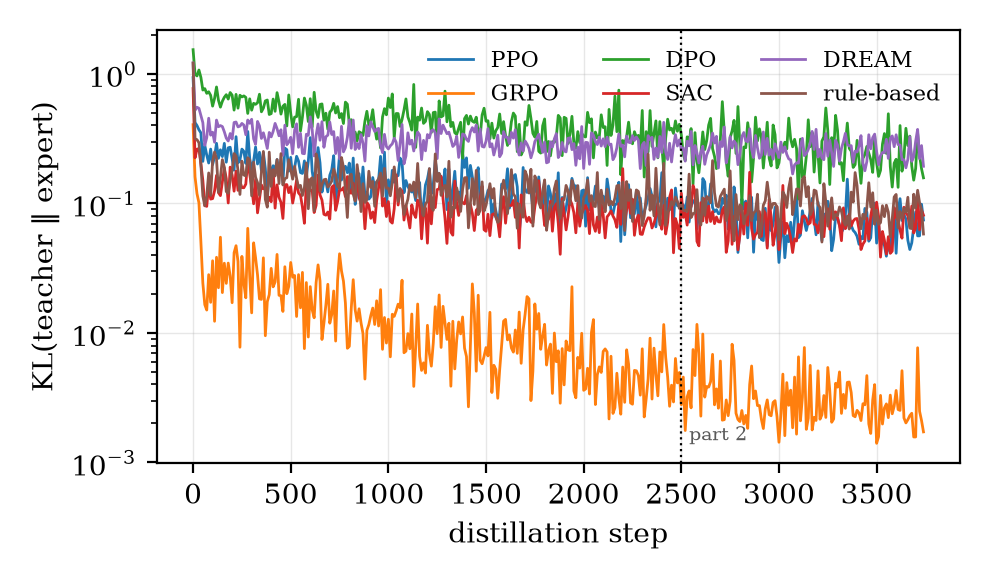}
\caption{Per-teacher distillation loss, the first term of
Eq.~\ref{eq:distill}, from the \texttt{A-plus} GRPO-league run, over
$2{,}500$ steps of part~1 (positions from teacher-vs-teacher games) and
$1{,}250$ steps of part~2 (positions from the student's own games, after the
dotted line). \texttt{A} and
\texttt{A-plus} distil from an identical teacher set and differ only in league
sampling, so this stage is the same in both; \texttt{A} was trained in an
earlier campaign and has no run logged alongside the ablation. Each expert head
learns from its own teacher at the same time. The DREAM and DPO experts plateau
highest (final-10\% KL $0.26$ and $0.24$); the rule-based expert, whose target
is a sampled empirical distribution rather than exact logits, ends at $0.09$,
close to SAC and PPO, and GRPO is matched almost exactly ($0.003$).}
\label{fig:distillkl}
\end{figure}

Figure~\ref{fig:distillkl} shows the per-teacher distillation losses behind
this, and Table~\ref{tab:moe} the routing that resulted.

With one seed per recipe and no recipe C (Section~\ref{sec:limits}), we
report this as a pattern that is consistent across four measurements, not as
an established effect.

\subsection{What the Mixture of Experts Contributes}\label{sec:moe}
Every agent in this paper has a six-expert policy head with a top-2 gate. The
load-balance curve in Section~\ref{sec:curves} shows the gate spreading its
choices over roughly $4.8$ to $5.8$ of its six experts in the GRPO league,
which is usually taken as a good sign. It does not show that the experts
matter: a gate can be perfectly balanced and still choose between experts that
all want the same move. So we measured directly whether the routing changes
the agent's moves.

\paragraph{Probe:} We replayed each recipe's best checkpoint against
\texttt{random} for 400 games, logging every state at which the policy was
queried (10{,}376 to 15{,}336 states per recipe). At each state we record the
gate weights, the action each individual expert would choose over the legal
set, and two counterfactuals:
\begin{itemize}
\item \textbf{Disagreement:} do the two selected experts want different
actions? If not, routing cannot matter.
\item \textbf{Flip rate:} does the top-2 mixture choose a different action than
the top-1 expert would on its own? This is the main measure: the fraction of
moves where using two experts instead of one actually changes the move.
\end{itemize}
We also measure whether routing is structured, as the total-variation distance
between the top-1 expert distribution within a game phase and the marginal
over all phases. Zero means the gate ignores the phase.

\begin{table}[h]\centering
\begin{tabular}{lrrrrrrr}
\toprule
Recipe & States & Eff.\ $E$ & $\bar w_1$ & Disagree & \textbf{Flip} & max TV & Dead \\
\midrule
\texttt{A} & 15{,}294 & 5.21 & 0.752 & 16.7\% & \textbf{2.72\%} & 0.58 & 0 \\
\texttt{A-plus} & 12{,}057 & 4.67 & 0.780 & 12.8\% & \textbf{1.82\%} & 0.50 & 1 \\
\texttt{B-grpo} & 10{,}376 & 4.78 & 0.675 & 5.6\% & \textbf{1.71\%} & 0.86 & 0 \\
\texttt{B-ppo} & 10{,}540 & 3.53 & 0.830 & 12.5\% & \textbf{1.33\%} & 0.39 & 0 \\
\texttt{D} & 15{,}336 & 4.03 & 0.677 & 5.7\% & \textbf{1.70\%} & 0.33 & 0 \\
\bottomrule
\end{tabular}

\caption{What the gate does, over 400 games per recipe on that recipe's best
checkpoint. $\bar w_1$ is the mean weight on the top-1 expert. ``Disagree'' is
the share of states where the two selected experts prefer different actions;
``Flip'' is the share where the mixture's action differs from the
top-1 expert's. ``max TV'' is the largest per-phase routing shift, and
``Dead'' counts experts never selected first.}
\label{tab:moe}
\end{table}

\paragraph{Routing depends on the game phase:} The gate does use its input.
Per-phase routing shifts reach a total variation of $0.86$ for \texttt{B-grpo}
and $0.33$--$0.58$ for the others, so which expert leads depends a lot on
whether the agent is drawing, responding to an \textsc{Attack}, or deciding in
a \textsc{Nope} window. \texttt{B-ppo} is the most concentrated, with one expert
leading $47\%$ of the time and an effective count of $3.53$ against
\texttt{A}'s $5.21$.

\paragraph{Routing rarely changes the move:} The flip rate is
$1.33\%$--$2.72\%$. For more than $97\%$ of moves, the whole mixture could be
replaced by the single expert the gate ranked first, and the move would be the
same. The disagreement column shows why: the two selected experts want the
same action $83\%$--$94\%$ of the time, and when they disagree, the mean top-1
weight of $0.68$--$0.83$ usually decides it in favour of the top expert. So the six expert heads, each with
$512 \times 186$ parameters, change about one move in fifty.

\paragraph{Differences between recipes:} \texttt{A}, distilled from five
different algorithms plus a heuristic, has the highest disagreement
($16.7\%$) and the highest flip rate ($2.72\%$); \texttt{B-grpo} and \texttt{D}, whose experts come from six
checkpoints of one algorithm and from random initialisation respectively, sit
at $5.6\%$ and $5.7\%$ disagreement. Some of the teacher diversity carries
over into the experts and can be measured here, but it is too small to change
many moves. This is the same result as Section~\ref{sec:distill}, seen from the
architecture side: the diversity was there after distillation and was mostly
lost during league play.

\paragraph{Caveats:} This probe plays against \texttt{random}, which leads to
different game states than league training does, and the effective expert
counts here ($3.53$--$5.21$) are lower than the
GRPO-league training-time load-balance figures ($4.8$--$5.8$) in
Section~\ref{sec:curves}. We also did not train a single-expert control at
matched parameter count, so we can say what the mixture does not add, but not
how a single head of the same size would have scored.

\subsection{Games Won vs. Matches Won}\label{sec:eval}
Table~\ref{tab:playoff} ranks the playoff by matches won instead of games
won.

\begin{table}[t]\centering
\begin{tabular}{rrlrrr}
\toprule
M\# & Elo\# & Entrant & W--L & Score\% & $\Delta$ \\
\midrule
1 & 1 & \texttt{B-grpo\_grpo\_g870000} & 19--0 & 54.5\% & +0 \\
2 & 3 & \texttt{B-grpo\_grpo\_g1120000} & 18--1 & 52.9\% & +1 \\
3 & 4 & \texttt{B-grpo\_grpo\_final} & 17--2 & 52.8\% & +1 \\
4 & 5 & \texttt{B-grpo\_grpo\_g1000000} & 15--4 & 52.6\% & +1 \\
5 & 6 & \texttt{B-grpo\_grpo\_g750000} & 15--4 & 52.5\% & +1 \\
6 & 7 & \texttt{B-grpo\_grpo\_g620000} & 13--6 & 51.8\% & +1 \\
7 & 10 & \texttt{B-grpo\_grpo\_g370000} & 10--9 & 50.1\% & +3 \\
8 & 11 & \texttt{D\_grpo\_g1120000} & 9--10 & 49.7\% & +3 \\
9 & 2 & \texttt{A\_grpo\_final} & 8--11 & 53.9\% & -7 \\
10 & 8 & \texttt{B-grpo\_grpo\_g500000} & 8--11 & 50.8\% & -2 \\
\bottomrule
\end{tabular}

\caption{Playoff top ten by match record. $\Delta$ is the shift from the
Bradley--Terry ranking. \champ{} is the only undefeated entrant; the
Bradley--Terry rank-2 agent, \texttt{A\_grpo\_final}, sits ninth with a
losing record.}
\label{tab:playoff}
\end{table}

Bradley--Terry fits individual games, so it rewards winning by large
margins. \texttt{A\_grpo\_final} wins its eight matches by 64.6\% on
average and loses its eleven by 46.1\%, which buys it more total games in the playoff (20{,}479 of 38{,}000) than every other entrant except the undefeated
\champ{} (20{,}716), which puts it second by Bradley--Terry. Ranked by matches
it falls seven places, to ninth.

\begin{table}[t]\centering
\begin{tabular}{rrlrrr}
\toprule
M\# & Elo\# & Entrant & W--L--T & Games\% & $\Delta$ \\
\midrule
1 & 2 & \texttt{B-grpo\_grpo\_g870000} & 47--2--0 & 57.0\% & +1 \\
2 & 3 & \texttt{B-grpo\_grpo\_final} & 46--3--0 & 56.1\% & +1 \\
3 & 9 & \texttt{B-grpo\_grpo\_g1120000} & 44--5--0 & 55.2\% & +6 \\
4 & 6 & \texttt{B-grpo\_grpo\_g1000000} & 43--5--1 & 55.5\% & +2 \\
5 & 8 & \texttt{B-grpo\_grpo\_g750000} & 43--6--0 & 55.2\% & +3 \\
6 & 10 & \texttt{B-grpo\_grpo\_g620000} & 40--8--1 & 54.9\% & +4 \\
7 & 12 & \texttt{B-grpo\_grpo\_g500000} & 40--9--0 & 53.7\% & +5 \\
8 & 1 & \texttt{A\_grpo\_final} & 38--11--0 & 58.5\% & -7 \\
9 & 4 & \texttt{A\_grpo\_g870000} & 37--12--0 & 56.1\% & -5 \\
10 & 11 & \texttt{D\_grpo\_g1120000} & 37--11--1 & 54.6\% & +1 \\
\bottomrule
\end{tabular}

\caption{The same reversal in the 50-agent super Elo, scored independently at
500 games per pair. \texttt{A\_grpo\_final} is Bradley--Terry rank 1 and
match-record rank 8, losing 11 of its 49 matches.}
\label{tab:super}
\end{table}

\begin{figure}[t]\centering
\includegraphics[width=0.49\textwidth]{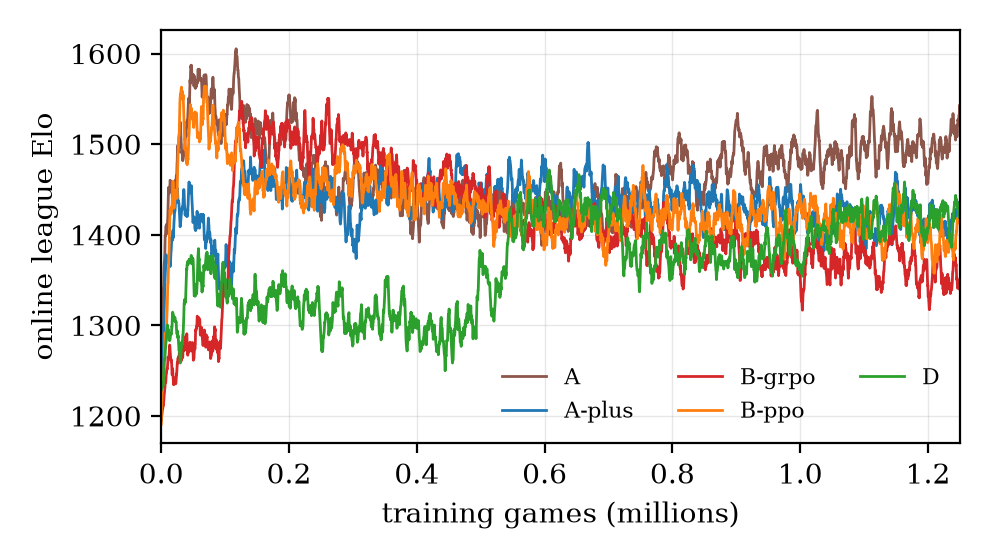}
\hfill
\includegraphics[width=0.49\textwidth]{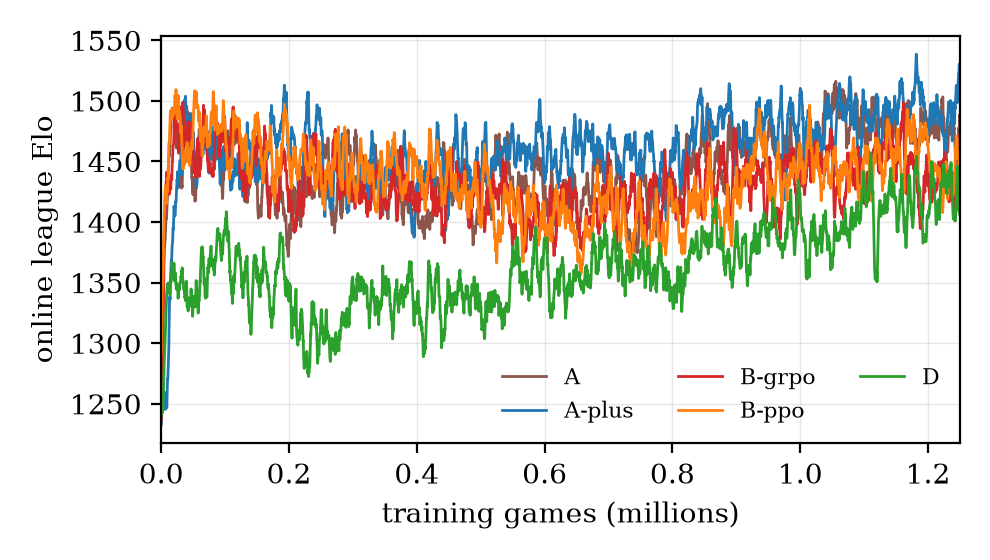}
\caption{Online league Elo: GRPO league (left), PPO league (right). This
rating is not used for any result in this paper; we show it because it does
not follow the offline Bradley--Terry ordering. Each run's rating is
computed inside that run against its own pool, so the curves are not
comparable across runs either.}
\label{fig:elo}
\end{figure}

Figure~\ref{fig:elo} shows why we do not use the league's own online rating
to settle this. The disagreement is not limited to one pair of agents. In the super Elo (the 50-agent round robin of Section~\ref{sec:protocol}),
in 15\% of the trios that can be drawn from the top 20, the three agents
beat each other in a circle (A beats B, B beats C, C beats A), so no ranking
can order all three correctly. In 26 of its 190 top-20 pairs, the lower-rated
agent wins the head-to-head clearly (its 95\% interval excludes 50\%). And 23 of
50 agents move two or more places depending on whether games or matches are
counted.
The same pattern appears in two separately fitted tournaments
(Tables~\ref{tab:playoff} and~\ref{tab:super}), so it does not depend on one
field of agents or one number of games per pair.

\paragraph{Against the LLMs:} Table~\ref{tab:llmagents} ranks the 20 playoff
agents by how often they beat the two LLMs, and Table~\ref{tab:rankcompare}
puts each agent's rank among the trained agents next to its results against
all four outside opponents. The three \texttt{A-plus} agents beat the LLMs
most often ($92.5$--$92.8\%$ over both), although they finish 16th, 17th and
20th in the playoff by matches won; the champion,
\texttt{B-grpo\_grpo\_g870000}, is fourth. Each LLM score comes from 200 games
per pair, a 95\% margin of about $\pm4$ points, so the top four agents are
statistically tied, but the gap between the top and the bottom ($92.8\%$
against $82.5\%$) is clear. Rank among the trained agents does not predict rank
against the LLMs: the Spearman correlation between the two is $-0.12$ for the
match-record rank and $-0.28$ for the Bradley--Terry rank.

\begin{table}[!htbp]\centering\small
\begin{tabular}{rlrrr}
\toprule
LLM\# & Agent & vs.\ Qwen3-30B & vs.\ Gemma-4-31B & Both \\
\midrule
1 & \texttt{A-plus\_grpo\_g1000000} & 92.5\% & 93.0\% & 92.8\% \\
1 & \texttt{A-plus\_grpo\_g870000} & 92.5\% & 93.0\% & 92.8\% \\
3 & \texttt{A-plus\_grpo\_g1120000} & 93.0\% & 92.0\% & 92.5\% \\
4 & \texttt{B-grpo\_grpo\_g870000} & 94.0\% & 90.5\% & 92.2\% \\
5 & \texttt{B-grpo\_grpo\_g750000} & 90.5\% & 89.5\% & 90.0\% \\
6 & \texttt{B-grpo\_grpo\_g250000} & 92.0\% & 86.5\% & 89.2\% \\
7 & \texttt{B-grpo\_grpo\_g620000} & 90.5\% & 87.5\% & 89.0\% \\
8 & \texttt{B-grpo\_grpo\_g500000} & 92.0\% & 85.0\% & 88.5\% \\
9 & \texttt{B-grpo\_grpo\_final} & 89.5\% & 86.5\% & 88.0\% \\
9 & \texttt{D\_grpo\_final} & 90.0\% & 86.0\% & 88.0\% \\
11 & \texttt{A\_grpo\_g750000} & 89.5\% & 85.0\% & 87.2\% \\
11 & \texttt{D\_grpo\_g1120000} & 83.0\% & 91.5\% & 87.2\% \\
13 & \texttt{B-ppo\_grpo\_g1000000} & 87.5\% & 86.5\% & 87.0\% \\
14 & \texttt{A\_grpo\_g870000} & 91.0\% & 82.5\% & 86.8\% \\
15 & \texttt{B-grpo\_grpo\_g1120000} & 87.0\% & 86.0\% & 86.5\% \\
16 & \texttt{B-grpo\_grpo\_g1000000} & 88.0\% & 84.5\% & 86.2\% \\
17 & \texttt{A\_grpo\_final} & 86.5\% & 85.5\% & 86.0\% \\
18 & \texttt{B-grpo\_grpo\_g370000} & 85.0\% & 86.5\% & 85.8\% \\
19 & \texttt{A\_grpo\_g1000000} & 88.5\% & 79.0\% & 83.8\% \\
20 & \texttt{A\_grpo\_g1120000} & 88.0\% & 77.0\% & 82.5\% \\
\bottomrule
\end{tabular}

\caption{The 20 playoff agents against the two LLMs: each agent's win rate over
200 games per LLM, sorted by the combined win rate over both (ties share a
rank). The 95\% margin on each entry is about $\pm4$ points, so neighbouring
ranks are not reliably different.}
\label{tab:llmagents}
\end{table}

\begin{table}[!htbp]\centering\small\setlength{\tabcolsep}{4pt}
\begin{tabular}{rrlrrrrr}
\toprule
M\# & Elo\# & Agent & W--L & LLM\# & vs.\ LLMs & vs.\ \texttt{random} & vs.\ \texttt{rule\_based} \\
\midrule
1 & 1 & \texttt{B-grpo\_grpo\_g870000} & 19--0 & 4 & 92.2\% & 89.3\% & 92.6\% \\
2 & 3 & \texttt{B-grpo\_grpo\_g1120000} & 18--1 & 15 & 86.5\% & 88.7\% & 93.1\% \\
3 & 4 & \texttt{B-grpo\_grpo\_final} & 17--2 & 9 & 88.0\% & 89.2\% & 94.3\% \\
4 & 5 & \texttt{B-grpo\_grpo\_g1000000} & 15--4 & 16 & 86.2\% & 89.6\% & 93.4\% \\
5 & 6 & \texttt{B-grpo\_grpo\_g750000} & 15--4 & 5 & 90.0\% & 88.6\% & 93.7\% \\
6 & 7 & \texttt{B-grpo\_grpo\_g620000} & 13--6 & 7 & 89.0\% & 90.5\% & 93.2\% \\
7 & 10 & \texttt{B-grpo\_grpo\_g370000} & 10--9 & 18 & 85.8\% & 90.5\% & 93.5\% \\
8 & 11 & \texttt{D\_grpo\_g1120000} & 9--10 & 11 & 87.2\% & 89.5\% & 94.9\% \\
9 & 2 & \texttt{A\_grpo\_final} & 8--11 & 17 & 86.0\% & 90.7\% & 94.9\% \\
10 & 8 & \texttt{B-grpo\_grpo\_g500000} & 8--11 & 8 & 88.5\% & 90.8\% & 92.8\% \\
11 & 14 & \texttt{B-grpo\_grpo\_g250000} & 8--11 & 6 & 89.2\% & 88.8\% & 92.0\% \\
12 & 15 & \texttt{D\_grpo\_final} & 8--11 & 9 & 88.0\% & 90.8\% & 94.6\% \\
13 & 12 & \texttt{A\_grpo\_g1120000} & 7--12 & 20 & 82.5\% & 87.1\% & 95.9\% \\
14 & 9 & \texttt{A\_grpo\_g870000} & 6--13 & 14 & 86.8\% & 88.8\% & 96.8\% \\
15 & 13 & \texttt{A\_grpo\_g1000000} & 6--13 & 19 & 83.8\% & 91.8\% & 95.1\% \\
16 & 17 & \texttt{A-plus\_grpo\_g1000000} & 6--13 & 1 & 92.8\% & 93.7\% & 97.0\% \\
17 & 19 & \texttt{A-plus\_grpo\_g1120000} & 5--14 & 3 & 92.5\% & 93.7\% & 97.9\% \\
18 & 16 & \texttt{B-ppo\_grpo\_g1000000} & 4--15 & 13 & 87.0\% & 85.9\% & 91.0\% \\
19 & 18 & \texttt{A\_grpo\_g750000} & 4--15 & 11 & 87.2\% & 90.1\% & 94.5\% \\
20 & 20 & \texttt{A-plus\_grpo\_g870000} & 4--15 & 1 & 92.8\% & 94.4\% & 97.1\% \\
\bottomrule
\end{tabular}

\caption{Each playoff agent's rank among the trained agents (M\#: by matches
won, with its win--loss record; Elo\#: by Bradley--Terry) next to its rank and
win rate against the two LLMs and its win rate against \texttt{random} and
\texttt{rule\_based} (1{,}000 games each). Sorted by M\#. The agents that rank
lowest among the trained agents include the ones that do best against the
LLMs.}
\label{tab:rankcompare}
\end{table}

\subsection{Specialists and Generalists}\label{sec:archetype}
Recipe A produces specialists. In the super Elo, \texttt{A\_grpo\_final} wins
$76\%$ of its games against its easiest opponent but only $37\%$ against its
hardest. B-grpo produces generalists: the champion, \champ{}, wins between
$49\%$ and $70\%$ against every opponent, loses only 2 of its 49 matchups in
the super Elo, and loses none of its 19 in the playoff.

This is why the two rankings disagree. When we rank the agents by matches won
instead of games won, A's four best checkpoints in the super Elo all move
down, by 5 to 8 places. B-grpo's four best all move up a little, by 1 to 3
places, and none of its GRPO-branch checkpoints moves down. The playoff shows
the same pattern for A, with smaller drops of 1 to 7 places.

\FloatBarrier
\section{Limitations}\label{sec:limits}
\textbf{One seed per recipe:} All ten runs used seed 42. Recipe-level comparisons
are single observations; the evaluation findings do not depend on them.
\textbf{The pending card is hidden:} In a \textsc{Nope} window the
observation reports only that an action is pending, not which of seven card
types it is. This affects $11.3\%$ of
decisions (2.83M of the 25.0M decisions in the playoff games). All agents and both LLMs
have the same handicap; it is a limitation of our observation encoder.
\textbf{No opponent history:} 96 observation dimensions reserved for
per-opponent play history are never written.
\textbf{Recipe C was cut:} Without a random-initialisation plus full-pool recipe, D
cannot tell whether distillation or the diverse pool made the difference.
\textbf{Small per-recipe $n$:} In Table~\ref{tab:axes} the field is the
playoff's top 20, itself selected by Elo.
\textbf{No single-head control for the MoE:} Section~\ref{sec:moe} shows what
the mixture does not add; without a single-expert head of the same size we
cannot say what it does add.
\textbf{Recipe A has no fixed-opponent logging:} The
\texttt{league/fixed\_opponent\_fraction} metric was added with the
\texttt{--min-fixed-fraction} flag, after recipe \texttt{A} had already
trained, so the baseline recipe is absent from Table~\ref{tab:fixedfrac} and
Figure~\ref{fig:fixedfrac}. The A-versus-A-plus outcome comparison in
Table~\ref{tab:axes} is unaffected, but the exposure explanation can be shown
only against \texttt{B} and \texttt{D}, which differ from A-plus in teachers
and pool as well as in sampling.
\textbf{Teachers only played the random opponent:} In Phase~1 every
specialist trained only against the uniform-random opponent. The code has a
pool for past checkpoints, but nothing was ever added to it, so the teachers
never played each other, their own earlier versions, or any stronger opponent
before distillation. Their Phase-1 win rates (Table~\ref{tab:phase1}) are
measured against the same opponent they trained against.
\textbf{Training never alternated seats:} In both phases the learner was
always seated as the first mover; only evaluation alternates. In evaluation, where seats
alternate, the first mover wins $63.4\%$ of the 612{,}500 super-Elo games
($+13.4$ pp), so every agent was
evaluated half the time from a seat it never trained in, and training-time
win rates are inflated by the seat. Because this holds identically for every
recipe, it does not by itself favour one recipe, but it may compress or reorder
the gaps between them, and second-seat play is likely under-trained across
the board.

\section{Conclusion}
League self-play produces agents that beat two open-weight LLMs, which see the
same information, by 6.5:1 to 8.5:1. Looking at the recipe in detail gives three
main results. Keeping fixed outside opponents
in training lowers an agent's rank against other league agents but makes it
the strongest recipe against every outside opponent we tested, so this setting
is a trade-off with no single best value. Distilling five different
specialists and a rule-based bot gave a student weaker than its best teacher,
and the choice of teacher algorithm mattered more than the number of different
teachers. The mixture-of-experts gate picks experts based on the game phase,
but using two experts instead of one changes the move in under 3\% of states,
so the gate being balanced does not mean the experts are useful.

There is also no single correct ranking of these agents: three reasonable
criteria pick three different best agents. We recommend reporting at least
three views (a rating, a match record, and results against outside opponents)
and stating which one any headline claim uses.

\bibliographystyle{plainnat}
\bibliography{references}
\end{document}